\documentclass[11pt]{article}

\usepackage[preprint]{acl}

\usepackage{times}
\usepackage{latexsym}
\usepackage[T1]{fontenc}

\usepackage[utf8]{inputenc}

\usepackage{microtype}

\usepackage{inconsolata}

\usepackage{graphicx}
\usepackage{amssymb}
\usepackage{amsmath}
\usepackage{booktabs}
\usepackage{tabularx}
\usepackage{url}
\usepackage{pgfplots}
\usepgfplotslibrary{groupplots}
\usetikzlibrary{calc}
\pgfplotsset{compat=1.18}
\title{Negation Beyond the Verbal Channel: Temporal Multimodal Correlates in Dialogue}

 \author{Leon Hammerla \and Patrick Schrottenbacher \and Alexander Mehler \\
        \texttt{\{hammerla $\cdot$ schrottenbacher $\cdot$ mehler\}@em.uni-frankfurt.de}\\
        Goethe University, Frankfurt am Main, Germany}

\begin{document}
\maketitle
\begin{abstract}
Negation is typically modeled through its linguistic realization, although spoken interaction is accompanied by tightly coordinated nonverbal behavior.
We ask whether contexts centered on spoken negation cues contain measurable multimodal behavioral information: whether they can be distinguished from matched control contexts without lexical or acoustic input, where this information occurs in time, which modalities carry it, and whether it extends to the dialogue partner.
We study 27 human-human interviews conducted in virtual reality, comprising temporally aligned gaze, facial, head, body, hand, and finger behavior and 964 annotated negation cues.
Treating classification as a predictive probe, we compare 20 time-series models while excluding lexical and acoustic information, and then systematically vary temporal context, interactional source, modality availability, and event timing.
Across grouped 10-fold cross-validation, the strongest probes reach up to .75 mean held-out AUROC from speaker-side behavior.
Temporal analyses show that predictive information is concentrated around cue onset but remains detectable over a broader surrounding interval, while dialogue-partner behavior carries weaker predictive information with a comparatively diffuse temporal profile.
Ablation and timing perturbations further show that facial features produce the largest modality-ablation effect and that the trained probe is sensitive to the temporal organization of the observed events.
\end{abstract}
\section{Introduction}
Human communication is inherently multimodal.
Spoken language is accompanied by gaze, facial behavior, head and body motion, and manual gestures, which constitute channels of communication that are closely coordinated in time~\citep{Goldin:Meadow:2013,Kelly:etal:2009}.
While computational models increasingly exploit such signals, it remains less clear how specific linguistic phenomena are reflected in nonverbal behavior, and at what temporal scale these effects become observable~\citep{Tsai:etal:2019}.
Negation provides a particularly interesting case. 
In computational linguistics, negation has predominantly been studied through its linguistic realization, including the identification of negation cues and their scope \citep{Morante:Blanco:2012}. 
At the same time, negative expressions can be systematically associated with nonverbal behavior, with head movements providing a particularly well-established example \citep{Kendon:2002}. 
Moreover, dialogue is a joint activity in which listeners continuously produce behavioral responses to speakers \citep{Clark:Krych:2004}. 
Explicit negation cues may therefore be associated with measurable patterns in the surrounding multimodal
interaction. 
Such patterns may be distributed across modalities, participants, and time rather than being tied to a single gesture or moment.
In this work, we study the temporal multimodal correlates of spoken negation cues in human-human dialogue.
We use temporally aligned multimodal interaction data and treat spoken words as anchors around which surrounding nonverbal events are observed. 
We then ask whether windows centered on negation cues can be distinguished from matched control-word windows using multimodal behavior without lexical or acoustic input.
We use classification as a predictive probe to quantify how cue-centered contexts differ from matched controls across temporal context, modality, and interactional source.
This provides a controlled way to investigate \emph{when}, \emph{where}, and \emph{through whom} behavioral information associated with explicit negation cues becomes predictive.
We address the following research questions:
\begin{itemize}
    \item \textbf{Discriminability:} Can negation-cue-centered contexts be distinguished from matched controls using only surrounding multimodal behavior?
    \item \textbf{Temporal extent:} How does this distinguishability change across different temporal contexts?
    \item \textbf{Modality contribution:} Which behavioral modalities contribute most strongly to cue-centered discriminability?
    \item \textbf{Interactional source:} How is cue-associated predictive information distributed between the speaker and the dialogue partner?
    \item \textbf{Temporal sensitivity:} Which behavioral event types show the greatest predictive sensitivity to their temporal relation to the spoken word?
\end{itemize}
By framing prediction as a probe of multimodal interaction, our study examines how explicit negation cues are embedded in coordinated, time-dependent human behavior rather than treating them only as isolated linguistic labels.
\section{Related Work}
\subsection{Negation in NLP}

Negation has been studied extensively in natural language processing, most commonly through the identification of \emph{negation cues} and the linguistic material falling within their \emph{scope}~\citep{Morante:Sporleder:2012,Morante:Blanco:2012}.
Annotated resources such as BioScope~\citep{Szarvas:etal:2008} and the \textsc{*SEM} 2012 shared task~\citep{Morante:Blanco:2012} established widely used formulations of cue, scope, and focus detection.
Early computational approaches included rule-based systems~\citep{Chapman:etal:2001,Peng:etal:2018} and feature-based statistical models that combined lexical and syntactic information with classifiers such as SVMs and CRFs ~\citep{Morante:Daelemans:2009,Lapponi:etal:2012}. 
Subsequent work increasingly adopted neural architectures~\citep{Fancellu:etal:2016}, including transformer-based systems such as \textsc{NegBERT}~\citep{Khandelwal:Sawant:2020} and syntax-aware graph attention extensions such as \textsc{D-Neg}~\citep{Hammerla:etal:2025}. 
Despite these advances, negation remains challenging for contemporary language models. 
Negated examples are comparatively underrepresented in common NLU benchmarks~\citep{Hossain:etal:2022}, pretrained language models can be insensitive to distinctions between affirmative and negated statements ~\citep{Kassner:Schuetze:2020}, and related limitations persist in larger autoregressive language models~\citep{Truong:etal:2023, GarciaFerrero:etal:2023}.
These approaches nevertheless treat negation primarily through linguistic input, leaving its manifestation in concurrent nonverbal behavior largely outside the modeling objective.
\subsection{Multimodal Negation}
Negation is not expressed exclusively through the verbal channel. 
Head shakes are among its best documented nonverbal correlates~\citep{Kendon:2002}, while manual gestures and combinations of head and hand movements have been shown to systematically accompany spoken negation and align with its node, scope, and focus~\citep{Harrison:2010,Harrison:2014,Harrison:Larrivee:2016}.
Prosodic and gestural cues can further affect the interpretation of negative
utterances~\citep{Prieto:etal:2013,Tubau:etal:2015,GonzalezFuente:etal:2015}, and gesture has been related to agreement and refusal~\citep{Guidetti:2005}, scope disambiguation~\citep{Brown:Kamiya:2019}, and covert negative meaning~\citep{Inbar:Shor:2019}; see \citet{Harrison:2024} for a recent overview.
Computational work has increasingly examined whether such distinctions are captured by multimodal models. Pretrained vision-language models have been shown to struggle with negation~\citep{Dobreva:Keller:2021}, motivating negation-focused multimodal evaluation~\citep{Parcalabescu:etal:2022,Sato:etal:2023} and negation-aware video retrieval~\citep{Wang:etal:2022}.
Subsequent work has analyzed how negation is represented within \textsc{CLIP}~\citep{Quantmeyer:etal:2024}, while more recent studies have developed dedicated benchmarks and training strategies for improving negation understanding in vision-language models~\citep{Park:etal:2025,Alhamoud:etal:2025}.
Most recently, \citet{Abusaleh:etal:2026} study cross-modal negation in video-text data through representation analysis and cross-modal attention fusion.
\subsection{Temporal Multimodal Interaction}
Nonverbal behavior is temporally coordinated with speech without necessarily being strictly synchronous, as shown for gesture-speech timing~\citep{Leonard:Cummins:2011} and for gaze coupling between speakers and
listeners~\citep{Richardson:Dale:2005,Richardson:etal:2007}.
Dialogue is further shaped by behavioral feedback from interlocutors, which speakers continuously monitor during interaction~\citep{Clark:Krych:2004}.
Computational work reflects this temporal and interactional perspective through multimodal corpora combining speech with gaze, gesture, and other behavioral signals~\citep{Carletta:etal:2006,Kontogiorgos:etal:2018,Kim:etal:2025}, as well as models designed to capture temporally unaligned cross-modal dependencies~\citep{Tsai:etal:2019}. 
These findings suggest that both the temporal context of a linguistic event and the source of the surrounding behavior may affect how much multimodal information it carries.
\paragraph{} Taken together, previous work establishes systematic links between spoken negation and nonverbal behavior, shows that multimodal models remain challenged by negation, and demonstrates that verbal and nonverbal behavior are temporally structured within dialogue.
However, the behavioral correlates surrounding spoken negation cues have not been systematically characterized from fine-grained, temporally aligned event streams spanning multiple modalities.
We address this gap using event logs from human-human interaction in virtual reality (VR), quantifying how cue-centered discriminability varies across temporal contexts, modalities, and interactional roles.
\section{Multimodal Dialogue Corpus}
\begin{figure}[ht]
\centering
\includegraphics[width=\linewidth,trim=8.5cm 1cm 9cm 8cm, clip]{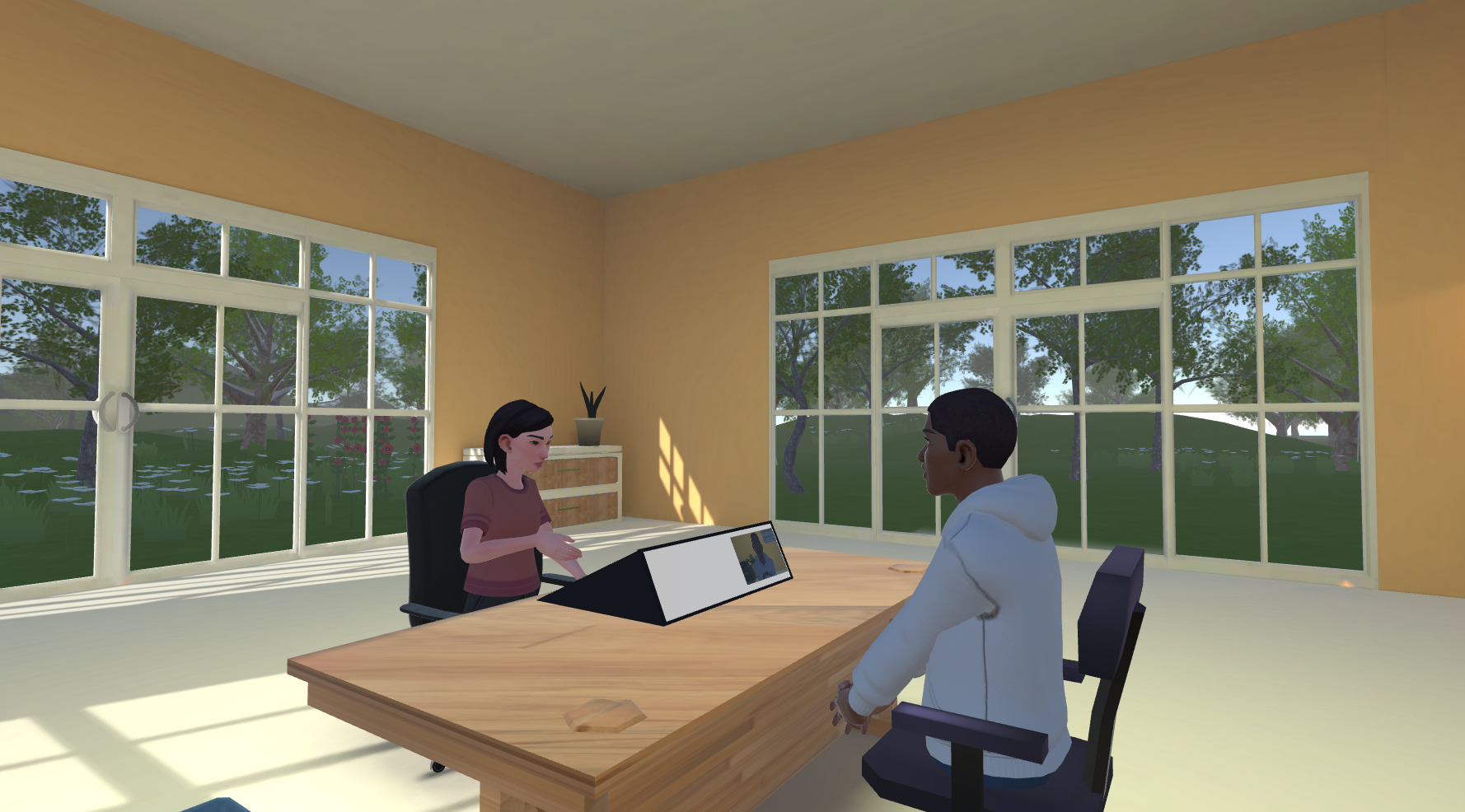}
\caption{
Interactional VR interview setting with the interviewer (left) navigating the survey questionnaire and the interviewee seated opposite.
}
\label{fig:scene}
\end{figure}
Our experiments use a German-language multimodal dialogue corpus from 27 individual survey interviews conducted in VR.\footnote{Project code and processed datasets are available at \url{https://github.com/vrneg/vrneg01} and \url{https://huggingface.co/VR-Faces-Neg}.
The dataset paper will be linked here once available.}
The recordings involve $30$ unique participants, with $3$ serving as interviewers and $27$ as interviewees.
During the study, participants were represented by avatars whose features were controlled through the Meta Quest Pro headsets they wore.
These headsets not only recorded their head and body positions within the room, but also their hands, fingers, gaze, face, and voice. 
All of these features were recorded at approximately $18.5$\,Hz throughout interviews lasting $27.7$ minutes on average.
The survey itself spanned five broad categories and purposefully included questions that were known to cause discomfort due to the sensitive nature of the topics (e.g. income and Machiavellianism).
A depiction of the interview environment and avatars can be seen in Figure~\ref{fig:scene}.
We organize the recordings in an event-based \textsc{SurrealDB} database~\citep{Surrealdb:2026}, with separate temporally aligned streams for body, eye, facial, head, hand, and finger events.
The audio recordings are stored in approximately $5$\,s chunks, while \textsc{CrisperWhisper (v1)}~\citep{Zusag:etal:2024} is applied to the full recordings to obtain word-level transcriptions and timestamps for alignment with the multimodal streams.
We annotate the resulting transcriptions for negation using \textsc{D-Neg}~\citep{Hammerla:etal:2025}, identifying both negation cues and their corresponding scopes (exact hyperparameter configurations for both models are provided in the Appendix~\ref{sec:neg-anno-conf}).
Across $57,971$ spoken words in the corpus, we identify $953$ negation instances, comprising $964$ cue words and $2,811$ scope words.
The cue distribution is skewed toward \textit{nicht}, which accounts for approximately 80\% of annotated cue instances.
The shared temporal representation enables extraction of multimodal windows around spoken words and locally matched controls from the surrounding conversational context.
\section{Task Formulation}
We operationalize the study of multimodal correlates surrounding explicit negation cues as a binary cue-centered classification problem.
Each spoken word serves as a temporal \emph{anchor}.
The positive class comprises words annotated as negation cues, whereas the negative class consists of words outside any annotated negation cue or scope.
Neither the lexical identity of the anchor word nor acoustic information is provided to the classifier; predictions are based exclusively on multimodal behavioral events surrounding the anchor.
We separately assess the potential contribution of visual articulation through a targeted lower-face ablation in Section~\ref{sec:modal}.
For an anchor word $w_i$ at time $t_i$, we define a temporal context by left and right window sizes $w_L$ and $w_R$ and collect
\[
E_i^{w_L,w_R}
=
\{e_j \mid t_i-w_L \leq t_j \leq t_i+w_R\}.
\]
The two window sizes can be varied independently, with symmetric windows satisfying $w_L = w_R$ and asymmetric windows allowing $w_L \neq w_R$.
Furthermore, setting a negative boundary ($w_L < 0$ or $w_R < 0$) creates an offset sliding window, effectively excluding the target event from the window itself.
Importantly, we do not define the context relative to the full word interval, i.e., from $w_L$ before word onset to $w_R$ after word offset.
Such a formulation would make the total observation interval dependent on the duration of the anchor word and could indirectly introduce lexical information, allowing the classifier to exploit systematic duration differences of frequent negation cues.
By default, event windows contain only behavior produced by the speaker of the anchor word.
To examine whether cue-associated predictive information extends beyond the speaker, we additionally consider a \emph{partner-only} condition containing events from the interlocutor and a \emph{dyadic} condition containing events from both participants.
These conditions allow us to test whether cue-associated predictive information is carried primarily by the speaker's own behavior or is also present in the behavior of the dialogue partner.
Together, these choices define two main dimensions of the task: the temporal context around the anchor, and the interactional source of the nonverbal events.
\section{Experimental Setup}
\subsection{Dataset Construction and Conditions}
We instantiate the task along the two main dimensions defined above:
(i) the temporal context, and
(ii) the interactional source.
For each of the 964 negation cue words (positive anchors), we sample one control word produced by the same speaker within the same $5$\,s audio chunk of the same interview, excluding all words annotated as negation cues or scopes. To avoid selecting temporally near-identical anchors, control words are required to occur more than $1$\,s from the corresponding negation cue. If no eligible control is available within the same audio chunk, we progressively expand the search to temporally adjacent chunks from the same speaker and interview until an eligible control is found. This yields a balanced dataset of 1,928 anchors while matching positive and negative examples as closely as possible in speaker identity and local conversational context.
We reuse the same positive anchors and sampled controls across all subsequent conditions.
For (i), we evaluate symmetric contexts with $w_L=w_R\in\{0.1,0.5,1,2.5\}$\,s, corresponding to total context spans of $0.2$, $1$, $2$, and $5$\,s, respectively.
Subsequent analyses explore asymmetric contexts by varying $w_L$ and $w_R$ independently. 
Furthermore, we employ sliding windows to localize cue-centered discriminability in time.
For (ii), each temporal condition is instantiated as \emph{speaker-only}, \emph{partner-only}, or \emph{dyadic}, according to the interactional source of the nonverbal events.
The same data partitions are retained across all conditions, enabling paired
comparisons across temporal contexts and interactional sources.
\subsection{Input Representations}
Each example contains the nonverbal events surrounding an anchor word across eight modalities: eye, facial, head, body, left/right hand, and left/right finger behavior.
Lexical identity, audio, absolute timestamps, participant identifiers, and database metadata are excluded.
For fixed-grid models, irregular event streams are converted into modality-specific numerical features, including gaze, facial-expression, head, body, hand, and finger-tracking measurements, together with derived linear and angular velocity features; a complete inventory of channels and their modality assignments is provided in Appendix~\ref{sec:feature-inventory}.
Continuous features are normalized using training data only and interpolated to 32 equidistant time steps, yielding $\mathbf{X}\in\mathbb{R}^{698\times32}$.
Missing positions are zero-filled, with presence indicators retained separately.
The event Transformer instead operates directly on the irregular event sequence using modality, interactional source, and relative timing.
High-frequency streams are limited to 128 observations per modality and window.
\subsection{Comparison Models}
We compare a broad set of time-series classifiers spanning several modeling paradigms (the exact model configurations are provided in the Appendix~\ref{sec:model-configurations}).
Random-convolution methods comprise \textsc{MiniRocket}~\citep{Dempster:etal:2021}, standard and sparse \textsc{MultiRocket+Hydra}~\citep{Tan:etal:2022,Dempster:etal:2023}, and \textsc{SelF-Rocket}~\citep{Lo:etal:2026}.
Feature-, dictionary-, shapelet-, and interval-based models include \textsc{Castor}~\citep{Samsten:etal:2024}, \textsc{Weasel 2.0}~\citep{Schfer:etal:2023}, \textsc{MrSQM}~\citep{Nguyen:Ifrim:2023}, and \textsc{DrCIF}~\citep{Middlehurst:etal:2021}, while \textsc{HIVE-COTE 2.0}~\citep{Middlehurst:etal:2021} provides an ensemble of complementary time-series classifiers.
We further evaluate three combinations of random-convolution features and pretrained tabular models: \textsc{RocketPFN}~\citep{Orourke:etal:2026}, \textsc{MASHT}~\cite{Cueppers:Vreeken:2026}, and adaptive \textsc{RocketPFN}.
Our neural comparison models comprise \textsc{Inception-TCN}~\citep{Bai:etal:2018,Ismail:etal:2020}, a compact fusion-\textsc{TCN}, and an irregular-event Transformer.
We additionally evaluate classification adaptations of \textsc{T2M-GPT}~\citep{Zhang:etal:2023}, a continuous-latent \textsc{T2M-GPT-v2} ablation of the same architecture, \textsc{MotionGPT}~\citep{Jiang:etal:2023}, the continuous diffusion-head mechanism of \textsc{MotionGPT3}~\citep{Zhu:etal:2026}, and the hierarchical short- and long-span \textsc{G-HTT} architecture~\citep{Wen:etal:2025}.
The comparison assesses whether the observed discriminability is robust across modeling approaches rather than introducing a new classification architecture.
\subsection{Cross-Validation and Evaluation}
We use stratified, group-based 10-fold cross-validation with VR experiments as groups, ensuring that no recording session is shared across training, validation, and test data.
In each rotation, one fold is used for testing, one for validation, and the
remaining eight for training.
All data-dependent preprocessing and model selection are performed without
access to the test fold.

Our primary metric is AUROC, which measures class distinguishability independently of a decision threshold; we additionally report macro-F$_1$.
Results are averaged over ten test folds with identical splits across conditions. We report bootstrap 95\% CIs over fold-wise scores, or fold-wise differences for paired contrasts; complete intervals are provided in the figures and tables.
\section{Results} 
We first compare classifiers to assess whether cue-centered discriminability is robust across modeling approaches, and then use one high-performing probe for the detailed analyses.
\input{Figures/model_comparison_figure_tikz}
\subsection{Model Comparison}
\label{sec:mcomp}
Across the four coarse temporal windows in the speaker-only condition, the strongest models consistently distinguish negation-cue-centered windows from matched control windows (Table~\ref{tab:model-comparison}; Figure~\ref{fig:model-comparison}).
\textsc{RocketPFN} achieves the highest mean AUROC (.728), followed closely by \textsc{HIVE-COTE~2.0} (.723) and \textsc{DrCIF} (.722).
Validation and test rankings are closely aligned (Table~\ref{tab:model-ranking}).
Macro-$F_1$ yields nearly the same ordering (Spearman $\rho=.97$), with \textsc{HIVE-COTE~2.0} (.666) and \textsc{RocketPFN} (.665) effectively tied at the top.
The larger neural sequence models, including the event Transformer and the motion-language adaptations, perform substantially worse than the strongest time-series probes and exhibit pronounced train-validation gaps, indicating overfitting under the available data regime.
Several models achieve comparable performance, indicating that cue-centered discriminability is not specific to a particular classifier. 
We use \textsc{RocketPFN} as a computationally efficient representative probe for the remaining analyses.
Across models, the $\pm0.1$\,s context performs worst, whereas the three wider windows yield similar AUROC (.660-.663).
This motivates the finer-grained temporal analysis in Section~\ref{sec:temporal}.
\subsection{Temporal Profile of Cue-Centered Discriminability}
\label{sec:temporal}
\begin{figure}[t]
\centering
\definecolor{cspk}{HTML}{08306B}
\definecolor{clis}{HTML}{4292C6}
\definecolor{cbth}{HTML}{969696}
\definecolor{metagrey}{HTML}{808080}
\begin{tikzpicture}
\begin{axis}[
  width=6.85cm, height=4.35cm, scale only axis,
  xmin=-0.45, xmax=15.45, ymin=0.470, ymax=0.800,
  xtick={0,1,2,3,4,5,6,7,8,9,10,11,12,13,14,15}, xticklabels={$-$5,$-$2.5,,$-$1.5,,$-$0.5,,$-$0.1,0.1,,0.5,,1.5,,2.5,5},
  xticklabel style={font=\scriptsize},
  ytick={0.50,0.55,0.60,0.65,0.70,0.75,0.80},
  yticklabels={$.50$,$.55$,$.60$,$.65$,$.70$,$.75$,$.80$},
  yticklabel style={font=\footnotesize},
  xlabel={window extent relative to cue onset (s)},
  ylabel={AUROC},
  xlabel style={font=\footnotesize, yshift=1pt},
  ylabel style={font=\footnotesize, yshift=-4pt},
  ymajorgrids, grid style={metagrey!25, line width=0.25pt},
  axis line style={metagrey!60, line width=0.4pt},
  tick style={metagrey!60, line width=0.4pt},
  legend style={at={(0.5,1.03)}, anchor=south, legend columns=3,
    font=\footnotesize, draw=none, fill=none, inner sep=1pt,
    /tikz/every even column/.append style={column sep=4pt}},
  legend image post style={scale=0.85},
]
\draw[metagrey, line width=0.5pt] (axis cs:7.5,0.470) -- (axis cs:7.5,0.800);
\node[font=\scriptsize, text=metagrey, anchor=north east, inner sep=2pt]
  at (axis cs:7.4,0.800) {before onset};
\node[font=\scriptsize, text=metagrey, anchor=north west, inner sep=2pt]
  at (axis cs:7.6,0.800) {after onset};
\addplot[metagrey, dashed, line width=0.4pt, forget plot, domain=-0.45:15.45,
  samples=2] {0.5};
\node[font=\scriptsize, text=metagrey, anchor=south west, inner sep=1.5pt]
  at (axis cs:0.05,0.472) {chance};

\addplot[fill=cspk, fill opacity=0.16, draw=none, forget plot]
  coordinates {(0,0.7779) (1,0.7772) (2,0.7678) (3,0.7616) (4,0.7500) (5,0.7432) (6,0.7272) (7,0.6984) (8,0.7120) (9,0.7368) (10,0.7527) (11,0.7566) (12,0.7706) (13,0.7701) (14,0.7660) (15,0.7327) (15,0.6608) (14,0.6805) (13,0.6828) (12,0.6852) (11,0.6769) (10,0.6751) (9,0.6575) (8,0.6422) (7,0.6189) (6,0.6525) (5,0.6680) (4,0.6866) (3,0.6890) (2,0.7003) (1,0.7043) (0,0.6886)} -- cycle;
\addplot[fill=clis, fill opacity=0.16, draw=none, forget plot]
  coordinates {(0,0.6442) (1,0.6478) (2,0.6560) (3,0.6506) (4,0.6354) (5,0.6372) (6,0.6409) (7,0.6275) (8,0.6368) (9,0.6390) (10,0.6237) (11,0.6375) (12,0.6473) (13,0.6478) (14,0.6500) (15,0.6697) (15,0.5888) (14,0.5745) (13,0.5836) (12,0.5791) (11,0.5653) (10,0.5674) (9,0.5833) (8,0.5712) (7,0.5576) (6,0.5723) (5,0.5788) (4,0.5701) (3,0.5832) (2,0.5739) (1,0.5752) (0,0.5778)} -- cycle;
\addplot[fill=cbth, fill opacity=0.16, draw=none, forget plot]
  coordinates {(0,0.5383) (1,0.5679) (2,0.5749) (3,0.5635) (4,0.5885) (5,0.5947) (6,0.6250) (7,0.6110) (8,0.6084) (9,0.6081) (10,0.6149) (11,0.6381) (12,0.6170) (13,0.6396) (14,0.6295) (15,0.6370) (15,0.5699) (14,0.5766) (13,0.5824) (12,0.5502) (11,0.5569) (10,0.5508) (9,0.5580) (8,0.5651) (7,0.5578) (6,0.5800) (5,0.5498) (4,0.5352) (3,0.4947) (2,0.5258) (1,0.5197) (0,0.4810)} -- cycle;
\addplot[cspk, line width=0.7pt, mark=*, mark size=1.25pt]
  coordinates {(0,0.7339) (1,0.7421) (2,0.7336) (3,0.7251) (4,0.7184) (5,0.7050) (6,0.6901) (7,0.6603) (8,0.6785) (9,0.6963) (10,0.7145) (11,0.7172) (12,0.7281) (13,0.7267) (14,0.7229) (15,0.6984)};
\addlegendentry{speaker}
\addplot[clis, line width=0.7pt, dashed, mark=square*, mark size=1.05pt]
  coordinates {(0,0.6160) (1,0.6168) (2,0.6215) (3,0.6188) (4,0.6065) (5,0.6113) (6,0.6124) (7,0.5959) (8,0.6073) (9,0.6162) (10,0.6004) (11,0.6063) (12,0.6169) (13,0.6195) (14,0.6170) (15,0.6337)};
\addlegendentry{listener}
\addplot[cbth, line width=0.7pt, dotted, mark=triangle*, mark size=1.35pt]
  coordinates {(0,0.5094) (1,0.5433) (2,0.5506) (3,0.5269) (4,0.5615) (5,0.5721) (6,0.6021) (7,0.5831) (8,0.5852) (9,0.5834) (10,0.5843) (11,0.5983) (12,0.5844) (13,0.6099) (14,0.6022) (15,0.6021)};
\addlegendentry{both}
\end{axis}
\end{tikzpicture}
\caption{Cue-centered discrimination AUROC across pre- and post-onset one-sided windows and interactional sources. Lines show means over ten folds; shading denotes bootstrap 95\% CIs.
}
\label{fig:temporal-source}
\end{figure}
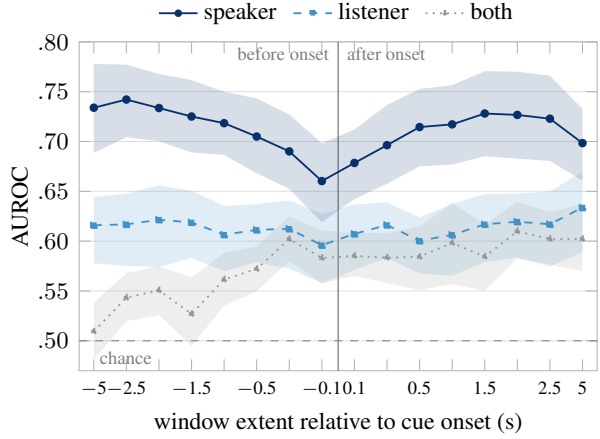
\begin{figure}[t]
\centering
\definecolor{cspk}{HTML}{08306B}
\definecolor{clis}{HTML}{4292C6}
\definecolor{cbth}{HTML}{969696}
\definecolor{metagrey}{HTML}{808080}

\begin{tikzpicture}

\begin{groupplot}[
    group style={
        group size=1 by 3,
        vertical sep=0.10cm
    },
    width=5.65cm,
    height=1.55cm,
    scale only axis,
    clip=false,
    xmin=-2.5,
    xmax=2.5,
    ymin=0.45,
    ymax=0.77,
    xtick={-2.5,-1.5,-0.5,0.5,1.5,2.5},
    ytick={0.50,0.60,0.70},
    yticklabels={$.50$,$.60$,$.70$},
    tick label style={font=\tiny},
    axis line style={metagrey!60, line width=0.4pt},
    tick style={metagrey!60, line width=0.4pt},
    ymajorgrids,
    grid style={metagrey!18, line width=0.25pt},
]

\nextgroupplot[
    xticklabels={},
]

\addplot[
    metagrey,
    dashed,
    line width=0.4pt,
    forget plot
] coordinates {
    (-2.5,0.5) (2.5,0.5)
};

\draw[
    metagrey!70,
    line width=0.5pt
]
(axis cs:0,0.45) -- (axis cs:0,0.77);

\node[
    rotate=-90,
    anchor=south,
    font=\footnotesize,
    text=cspk
] at (axis description cs:1.01,0.5) {\textbf{speaker}};

\addplot+[
    only marks,
    color=cspk,
    mark=*,
    mark size=1.7pt,
    error bars/.cd,
    y dir=both,
    y explicit,
    error bar style={line width=0.55pt},
    error mark options={rotate=90, mark size=1.5pt}
]
table[
    x=x,
    y=y,
    y error plus=ep,
    y error minus=em
] {
x       y       ep      em
-2.25   0.6475  0.0450  0.0424
-1.75   0.6544  0.0424  0.0402
-1.25   0.6719  0.0391  0.0398
-0.75   0.6630  0.0410  0.0399
-0.25   0.7050  0.0381  0.0365
 0.25   0.7145  0.0392  0.0393
 0.75   0.6925  0.0488  0.0485
 1.25   0.6614  0.0363  0.0338
 1.75   0.6560  0.0345  0.0337
 2.25   0.6385  0.0432  0.0409
};

\nextgroupplot[
    xticklabels={},
    ylabel={AUROC},
    ylabel style={font=\scriptsize, yshift=-3pt},
]

\addplot[
    metagrey,
    dashed,
    line width=0.4pt,
    forget plot
] coordinates {
    (-2.5,0.5) (2.5,0.5)
};

\draw[
    metagrey!70,
    line width=0.5pt
]
(axis cs:0,0.45) -- (axis cs:0,0.77);

\node[
    rotate=-90,
    anchor=south,
    font=\footnotesize,
    text=clis
] at (axis description cs:1.01,0.5) {\textbf{listener}};

\addplot+[
    only marks,
    color=clis,
    mark=square*,
    mark size=1.55pt,
    error bars/.cd,
    y dir=both,
    y explicit,
    error bar style={line width=0.55pt},
    error mark options={rotate=90, mark size=1.5pt}
]
table[
    x=x,
    y=y,
    y error plus=ep,
    y error minus=em
] {
x       y       ep      em
-2.25   0.6222  0.0218  0.0241
-1.75   0.6157  0.0262  0.0270
-1.25   0.6101  0.0274  0.0256
-0.75   0.6172  0.0231  0.0246
-0.25   0.6113  0.0259  0.0312
 0.25   0.6004  0.0224  0.0312
 0.75   0.6171  0.0289  0.0337
 1.25   0.6022  0.0271  0.0376
 1.75   0.5908  0.0302  0.0353
 2.25   0.5952  0.0330  0.0429
};

\nextgroupplot[
    xlabel={time relative to cue onset (s)},
    xlabel style={font=\scriptsize, yshift=1pt},
]

\addplot[
    metagrey,
    dashed,
    line width=0.4pt,
    forget plot
] coordinates {
    (-2.5,0.5) (2.5,0.5)
};

\draw[
    metagrey!70,
    line width=0.5pt
]
(axis cs:0,0.45) -- (axis cs:0,0.77);

\node[
    rotate=-90,
    anchor=south,
    font=\footnotesize,
    text=cbth
] at (axis description cs:1.01,0.5) {\textbf{both}};

\addplot+[
    only marks,
    color=cbth,
    mark=triangle*,
    mark size=1.8pt,
    error bars/.cd,
    y dir=both,
    y explicit,
    error bar style={line width=0.55pt},
    error mark options={rotate=90, mark size=1.5pt}
]
table[
    x=x,
    y=y,
    y error plus=ep,
    y error minus=em
] {
x       y       ep      em
-2.25   0.4965  0.0405  0.0402
-1.75   0.5074  0.0365  0.0413
-1.25   0.5431  0.0331  0.0426
-0.75   0.5140  0.0265  0.0281
-0.25   0.5721  0.0227  0.0232
 0.25   0.5843  0.0308  0.0329
 0.75   0.5899  0.0403  0.0417
 1.25   0.5574  0.0301  0.0301
 1.75   0.5470  0.0268  0.0261
 2.25   0.5437  0.0265  0.0227
};

\end{groupplot}
\end{tikzpicture}

\caption{AUROC in non-overlapping 500,ms windows around cue onset across interactional sources. Whiskers denote bootstrap 95\% CIs.}
\label{fig:sliding-window}

\end{figure}
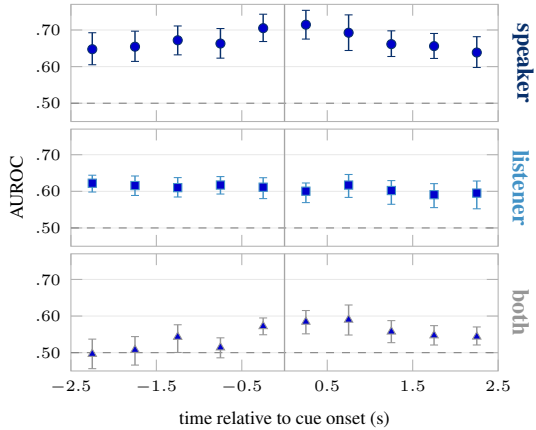
We next examine how discriminability changes with temporal context across speaker-only, partner-only, and dyadic conditions.
We focus here on the speaker temporal profile and analyze differences between interactional sources in Section~\ref{sec:interactional}.
To characterize where cue-associated predictive information occurs relative to cue onset, we re-evaluate \textsc{RocketPFN} on one-sided windows with extents $t\in\{.1,.25,.5,1,1.5,2,2.5,5\}$\,s, taken either before cue onset, $[-t,0]$, or after it, $[0,t]$ (Table~\ref{tab:temporal-source}; Figure~\ref{fig:temporal-source}).
Because larger windows cumulatively add events, the curves reflect the amount of usable evidence available up to each extent rather than the contribution of individual temporal segments.
For speaker behavior, discriminability is already above chance before cue onset: AUROC reaches $.660$ (95\% CI $[.619,.698]$) for $[-0.1,0]$\,s and increases to $.742$ (95\% CI $[.704,.777]$) at $2.5$\,s.
Thus, behavior preceding the spoken cue alone is sufficient to distinguish cue-centered from matched control contexts.
Post-onset speaker windows show a similar pattern, rising from $.678$ to a maximum of approximately $.73$.
Matched pre- and post-onset windows differ only modestly across temporal extents (paired $\Delta\mathrm{AUROC}$ ranging from $-.018$ to $.035$), with intervals including zero except at $t=5$\,s, where a pre-onset advantage emerges ($\Delta\mathrm{AUROC}=.035$, 95\% CI $[.010,.061]$). Performance otherwise largely saturates between $1.5$ and $2.5$\,s.
Neither side improves further when extending the window to $5$\,s.
To separate temporal location from the amount of accumulated evidence, we re-run the probe on ten disjoint $0.5$\,s windows tiling $[-2.5,2.5]$\,s (Figure~\ref{fig:sliding-window}).
Speaker performance peaks in the two windows adjacent to cue onset and declines steadily with temporal distance ($\rho=-.96$), from approximately $.710$ near onset to $.643$ at $\pm2.25$\,s.
Nevertheless, the bootstrap 95\% confidence interval lies entirely above $\mathrm{AUROC}=.5$ in every window, including $[-2.5,-2]$\,s (AUROC $.648$, 95\% CI $[.604,.691]$), showing that the pre-cue discriminability is not solely driven by behavior immediately adjacent to cue onset.
The two onset-adjacent windows show similar discriminability, with $\Delta\mathrm{AUROC}=.010$ for $[0,.5]$\,s relative to $[-.5,0]$\,s (95\% CI $[-.007,.024]$), indicating that performance is strongest around cue onset without a clear pre/post asymmetry at this temporal resolution.
\section{Analysis} 
Having established cue-centered discriminability and its temporal profile, we next assess its robustness to stricter control matching and investigate its interactional source, modality contributions, and temporal sensitivity.
\subsection{Robustness to Alternative Control Matching}
Our primary controls match speaker identity and local conversational context but may differ from negation cues in grammatical category or turn position.
We therefore construct stricter controls additionally matched by part of speech and turn-relative position, relaxing the same-chunk requirement and retaining 830 of 964 cues. 
On the same anchors, we also evaluate weaker controls sampled randomly outside negation cues and scopes.
At the symmetric $\pm1$\,s context, \textsc{RocketPFN} reaches mean held-out AUROC of \(0.723\), \(0.712\), and \(0.722\) under stricter, default, and weaker controls, respectively. 
Relative to the default condition, neither stricter matching ($\Delta$AUROC \(=+0.012\), 95\% CI \([-0.007,0.030]\)) nor weaker matching (\(+0.010\), \([-0.019,0.038]\)) reliably changes performance, indicating that additionally matching part of speech and turn position does not substantially reduce discriminability.
We next test whether discriminability primarily reflects the broader context of a negated expression. 
We sample negative anchors from within the annotated scope of the corresponding cue, requiring at least \(1.1\)~s separation from cue onset. 
This yields 311 cues. 
On exactly these anchors, performance drops from \(0.730\) with the default controls to \(0.677\) with within-scope controls ($\Delta$AUROC \(=-0.053\), 95\% CI \([-0.096,-0.012]\)).
Thus, matching within the same negation scope reduces but does not eliminate discriminability: cue-centered behavior remains distinguishable from behavior centered on another word within the same scope.
\subsection{Speaker and Dialogue-Partner Contributions}
\label{sec:interactional}
We first compare speaker-only, partner-only, and dyadic contexts to determine how predictive information is distributed across participants (Figures~\ref{fig:temporal-source} and~\ref{fig:sliding-window}).
Across the cumulative windows, speaker behavior is consistently more informative than partner behavior, averaging AUROC $.712$ versus $.614$ (paired $\Delta\mathrm{AUROC}=.098$, 95\% CI $[.063,.139]$), with higher speaker performance in all 16 temporal conditions.
Partner-only performance nevertheless yields mean AUROC above $.5$ throughout, with a minimum of $.596$ (95\% CI $[.558,.628]$), indicating that the interlocutor's behavior alone carries information predictive of whether the speaker's anchor is a negation cue.
The sliding-window analysis reveals that the partner temporal profile differs qualitatively from the speaker profile.
While speaker performance peaks around cue onset, partner performance remains nearly constant across $[-2.5,2.5]$\,s ($\rho=-.07$; AUROC $.591$-$.622$), with mean AUROC exceeding $.5$ in all ten windows.
Thus, the partner temporal profile is more consistent with broader interactional or contextual information associated with cue occurrence than with a response time-locked to the cue.
The dyadic condition performs worst in both analyses (AUROC $.575$ cumulative; $.546$ sliding-window). 
This likely reflects the fixed-grid representation, which averages participants and removes actor identity.
\subsection{Modality Contributions}
\label{sec:modal}
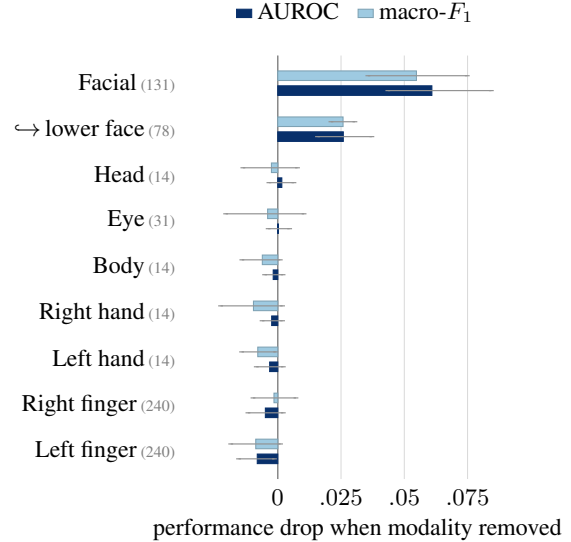
\begin{figure}[t]
\centering
\definecolor{barA}{HTML}{08306B}
\definecolor{barF}{HTML}{9ECAE1}
\definecolor{metagrey}{HTML}{808080}
\begin{tikzpicture}
\begin{axis}[
  xbar, y dir=reverse,
  width=4.25cm, height=5.60cm, scale only axis,
  bar width=3.6pt,
  enlarge y limits=0.075,
  xmin=-0.032, xmax=0.095,
  symbolic y coords={m0,m1,m2,m3,m4,m5,m6,m7,m8}, ytick=data,
  yticklabels={Facial\,\textcolor{metagrey}{\tiny(131)},$\hookrightarrow$\,lower face\,\textcolor{metagrey}{\tiny(78)},Head\,\textcolor{metagrey}{\tiny(14)},Eye\,\textcolor{metagrey}{\tiny(31)},Body\,\textcolor{metagrey}{\tiny(14)},Right hand\,\textcolor{metagrey}{\tiny(14)},Left hand\,\textcolor{metagrey}{\tiny(14)},Right finger\,\textcolor{metagrey}{\tiny(240)},Left finger\,\textcolor{metagrey}{\tiny(240)}},
  yticklabel style={font=\footnotesize, align=right},
  scaled x ticks=false,
  xtick={0,0.025,0.05,0.075}, xticklabels={$0$,$.025$,$.05$,$.075$},
  xticklabel style={font=\footnotesize},
  xlabel={performance drop when modality removed},
  xlabel style={font=\footnotesize, yshift=2pt},
  xmajorgrids, grid style={metagrey!30, line width=0.25pt},
  extra x ticks={0}, extra x tick labels={},
  extra x tick style={grid=major,
    grid style={metagrey, line width=0.5pt}},
  axis line style={draw=none}, tick style={draw=none},
  every axis plot/.append style={
    error bars/x dir=both, error bars/x explicit,
    error bars/error bar style={metagrey!85, line width=0.45pt},
    error bars/error mark options={metagrey!85, mark size=0.9pt, line width=0.45pt}},
  legend style={at={(0.5,1.055)}, anchor=south, legend columns=2,
    font=\footnotesize, draw=none, fill=none,
    inner sep=1pt, /tikz/every even column/.append style={column sep=5pt}},
  legend image code/.code={\filldraw[#1] (0cm,-0.055cm) rectangle (0.2cm,0.055cm);},
]
\addplot[fill=barA, draw=barA] coordinates {
  (0.0609,m0) += (0.0235,0) -= (0.0174,0)
  (0.0259,m1) += (0.0113,0) -= (0.0102,0)
  (0.0016,m2) += (0.0048,0) -= (0.0051,0)
  (0.0002,m3) += (0.0045,0) -= (0.0040,0)
  (-0.0018,m4) += (0.0039,0) -= (0.0034,0)
  (-0.0024,m5) += (0.0043,0) -= (0.0038,0)
  (-0.0032,m6) += (0.0054,0) -= (0.0053,0)
  (-0.0049,m7) += (0.0071,0) -= (0.0069,0)
  (-0.0081,m8) += (0.0068,0) -= (0.0074,0)
};
\addlegendentry{AUROC}
\addplot[fill=barF, draw=barF!80!black] coordinates {
  (0.0548,m0) += (0.0201,0) -= (0.0192,0)
  (0.0258,m1) += (0.0047,0) -= (0.0048,0)
  (-0.0025,m2) += (0.0103,0) -= (0.0113,0)
  (-0.0040,m3) += (0.0144,0) -= (0.0166,0)
  (-0.0061,m4) += (0.0071,0) -= (0.0081,0)
  (-0.0096,m5) += (0.0115,0) -= (0.0130,0)
  (-0.0079,m6) += (0.0071,0) -= (0.0064,0)
  (-0.0015,m7) += (0.0087,0) -= (0.0083,0)
  (-0.0087,m8) += (0.0098,0) -= (0.0099,0)
};
\addlegendentry{macro-$F_1$}
\end{axis}
\end{tikzpicture}
\caption{Leave-one-modality-out performance decrease at $w_L=w_R=1$,s. The indented lower-face row is a nested facial ablation; whiskers denote bootstrap 95\% CIs.
}
\label{fig:modality-ablation}
\end{figure}
For the modality and temporal-robustness analyses, we fix the context to $w_L=w_R=1$\,s, an intermediate window within the performance plateau observed above.
We estimate modality-specific predictive contributions using leave-one-modality-out ablation. For each modality, we retrain and evaluate the \textsc{RocketPFN} probe with that modality excluded from both training and test inputs, while retaining identical cross-validation splits across conditions (Figure~\ref{fig:modality-ablation}).
We define $\Delta m=m_{\mathrm{full}}-m_{\mathrm{ablated}}$; positive values therefore indicate performance lost when a modality is unavailable, rather than causal necessity or uniquely attributable information.
The full-input probe reaches AUROC $.734$ and macro-$F_1$ $.667$.
Facial behavior is the only modality whose removal reliably degrades the probe ($\Delta\mathrm{AUROC}=.061$, 95\% CI $[.044,.084]$; $\Delta$macro-$F_1=.055$).
To assess whether this effect is driven primarily by visual articulation of the cue word, we additionally remove the mouth-, lip-, chin-, cheek-, and jaw-related lower-face subset defined in Appendix~\ref{sec:feature-inventory}. 
This produces a smaller but reliable performance decrease ($\Delta\mathrm{AUROC}=.026$, 95\% CI $[.016,.037]$).
Thus, lower-face behavior contributes predictive information, but the selected articulatory features account for only part of the facial contribution and only part of overall cue-centered discriminability: substantial discriminability remains when lower-face features are unavailable, and even when the entire facial modality is removed.
No other modality shifts AUROC by more than $.009$, and all remaining intervals overlap zero except left-finger ablation, which marginally improves performance (95\% CI $[-.016,-.001]$), consistent with the sparse coverage of the finger streams ($60$-$65\%$ of windows absent).
\subsection{Temporal Event Robustness}
\hyphenation{timing-de-pen-dent single-mo-dal-ity right-hand}
\label{sec:robustness}
\input{Figures/dtw_robustness_figure_tikz}
The ablation analysis identifies which modalities contribute predictive information, but not whether the trained probe is sensitive to their temporal organization.
We therefore perturb the frozen \textsc{RocketPFN} checkpoints at inference time using rigid shifts of $\pm2$ grid steps ($\pm.129$\,s) and temporal rescaling by $.8$ or $1.2$ around the window centre.
Warps are applied either jointly to all modalities or to one modality in isolation (Figure~\ref{fig:dtw-robustness}).
Joint warping reduces AUROC by $.030$-$.051$ from the $.734$ baseline.
Rescaling is more damaging than rigid shifting ($\Delta\mathrm{AUROC}=.009$, 95\% CI $[.001,.015]$), and dilation more damaging than compression ($.020$, 95\% CI $[.008,.033]$), indicating sensitivity to changes in temporal organization.
Single-modality effects are smaller and less consistent, led by facial ($\Delta\mathrm{AUROC}=.023$) and right-hand ($.009$) timing, and are therefore treated as descriptive.
\section{Conclusion}
We investigated whether and to what extent explicit spoken negation cues are associated with distinguishable patterns of multimodal behavior during human-human dialogue.
Using lexical and acoustic information only to define temporal anchors, we treated multimodal classification as a predictive probe of gaze, facial, head, body, hand, and finger behavior surrounding negation cues. 
For speakers, strong statistical time-series models reliably distinguish negation-cue-centered from matched control contexts using multimodal behavior without lexical or acoustic input, including under stricter POS- and turn-position-matched controls, reaching up to .750 fold-mean test AUROC. 
Under the harder within-scope control, performance decreases from .730 to .677 on matched anchors ($\Delta$AUROC \(=-.053\), 95\% CI \([-.096,-.012]\)), indicating that broader negation context contributes to, but does not fully account for, cue-centered discriminability.
Increasing the available context improves discriminability up to approximately $2.5$\,s before and $1.5$\,s after negation cue onset, while the sliding-window analysis shows that discriminability is highest in the 0.5\,s immediately before and after the cue.
Dialogue-partner behavior is also predictive, reaching .634 AUROC, but remains nearly constant across both cumulative temporal extents and disjoint sliding windows, consistent with diffuse interactional or contextual information rather than a response tightly locked to the negation cue.
Facial features produce the largest modality-ablation effect, but removing only mouth-, lip-, chin-, cheek-, and jaw-related lower-face channels reduces AUROC by just $.026$ (from $.734$ to approximately $.708$), showing that visual articulation contributes to the signal but does not account for it: substantial predictive information remains without these articulatory features.
Finally, perturbing event timing reduces AUROC by .030-.051, showing that the trained probe is sensitive to changes in the temporal organization of its inputs.
Together, these findings show that explicit negation cues in this dialogue setting are embedded in distinguishable and temporally structured patterns of multimodal behavior, characterized by cue-localized speaker information, weaker and temporally diffuse partner information, and the largest modality-ablation effect for facial features.
\section*{Limitations}
Our findings should be interpreted within several constraints.
The study is based on 27 human-human interviews conducted in a single VR survey setting, so it remains unclear to what extent the observed behavioral patterns generalize to face-to-face interaction, other conversational tasks, populations, or languages. 
In addition, negation cues and their word-level timestamps are obtained automatically using \textsc{D-Neg} and \textsc{CrisperWhisper}; residual annotation or alignment errors may therefore affect particularly fine-grained temporal analyses. 
Our formulation focuses on explicitly marked negation cues and does not cover the broader space of implicit negative meaning, disagreement, refusal, or other pragmatically negative constructions.
Finally, the cue inventory is dominated by \textit{nicht} ($\approx 80\%$ of cue instances).
Our results therefore primarily characterize explicit negation cues under the lexical distribution of this corpus, where \textit{nicht} is the predominant realization, and do not establish invariance across different lexical realizations of negation.
\section*{Ethical Considerations}
The underlying VR recordings contain potentially sensitive behavioral data, including facial, gaze, and body-motion signals. 
We therefore do not release the original continuous recordings or metadata that would allow observations to be linked back to individual participants or recording sessions.
The released research data are restricted to the multimodal event windows used for the negation experiments and are stripped of participant identifiers, experiment identifiers, absolute timestamps, and other linkage metadata.
Consequently, individual examples cannot be directly associated with a particular participant, interview, or position within an interview from the released dataset.
This data-minimization strategy is intended to reduce privacy and re-identification risks while retaining the information necessary to reproduce the analyses reported in this work.
%
%
%
\bibliography{custom,anthology-1,anthology-2}
\appendix
\section{Annotation Model Configurations}
\label{sec:neg-anno-conf}
\paragraph{\textsc{CrisperWhisper}}
We use \path{nyrahealth/CrisperWhisper} with German as the decoding language and beam search with five beams.
The resulting word-level timestamps are retained for alignment with the multimodal event stream.
\paragraph{\textsc{D-Neg}}
Negation cues and scopes are annotated using \textsc{D-Neg} v0.1.1~\citep{Hammerla:etal:2025} with the German syntax-aware configuration.
We use the \path{D-NEG/cue-gat-de-sfu} and \path{D-NEG/scope-gat-de-sfu} checkpoints for cue detection and scope resolution, respectively, with a maximum sequence length of 256.
German syntactic annotations are produced with spaCy's \path{de_core_news_sm} model v3.8.0.
All remaining inference parameters use the respective package defaults.
\section{Main Model Configurations}
\label{sec:model-configurations}
\subsection{Multimodal Feature Inventory}
\label{sec:feature-inventory}
\begin{table*}[!t]
\centering
\small
\begin{tabular}{@{}lrrrrrr@{}}
\toprule
Modality & Recorded & Derived motion & Finger flags & Presence & Total & Global indices \\
\midrule
Eye          & 18  & 12 & 0  & 1 & 31  & $0$--$30$     \\
Facial       & 67  & 63 & 0  & 1 & 131 & $31$--$161$   \\
Head         & 7   & 6  & 0  & 1 & 14  & $162$--$175$  \\
Body         & 7   & 6  & 0  & 1 & 14  & $176$--$189$  \\
Left hand    & 7   & 6  & 0  & 1 & 14  & $190$--$203$  \\
Right hand   & 7   & 6  & 0  & 1 & 14  & $204$--$217$  \\
Left finger  & 130 & 96 & 13 & 1 & 240 & $218$--$457$  \\
Right finger & 130 & 96 & 13 & 1 & 240 & $458$--$697$  \\
\midrule
Total        & 373 & 291 & 26 & 8 & 698 & $0$--$697$    \\
\bottomrule
\end{tabular}
\caption{Fixed-grid channel allocation.
Global indices are zero-based positions along the first axis of
$\mathbf{X}$.
Recorded dimensions contain gaze measurements, facial blendshape weights and
tracking metadata, rigid-body poses, and finger-tracking measurements.
Derived dimensions contain first-order linear, angular, blendshape, bone,
and pinch motion features as applicable.
The eye modality corresponds to the dedicated gaze-tracking stream; ocular
expression blendshapes supplied by the facial tracker remain part of the
facial modality.}
\label{tab:fixed-grid-channels}
\end{table*}
\begin{table*}[!t]
\centering
\footnotesize
\begin{tabular}{@{}p{0.18\textwidth}p{0.20\textwidth}p{0.55\textwidth}@{}}
\toprule
Modality & Local channel index or name & Feature mapping \\
\midrule
Eye
& $0$; $9$
& Left- and right-eye gaze confidence, respectively. \\

Eye
& $1$; $10$
& Left- and right-eye gaze-validity indicators. \\

Eye
& $2$--$4$; $11$--$13$
& Left- and right-eye gaze positions $(x,y,z)$. \\

Eye
& $5$--$8$; $14$--$17$
& Left- and right-eye gaze orientations as quaternions $(x,y,z,w)$. \\

Eye
& $18$--$20$; $24$--$26$
& Left- and right-eye linear velocities $(v_x,v_y,v_z)$. \\

Eye
& $21$--$23$; $27$--$29$
& Left- and right-eye angular velocities
$(\omega_x,\omega_y,\omega_z)$. \\
\midrule
Facial
& $0$--$62$
& The 63 facial expression weights enumerated in
Table~\ref{tab:facial-feature-mapping}. \\

Facial
& $63$--$64$
& The two entries of the recorded
\texttt{expressionWeightConfidences} array, in stored order. \\

Facial
& $65$
& The facial-tracking \texttt{status.IsValid} indicator. \\

Facial
& $66$
& The \texttt{status.IsEyeFollowingBlendshapesValid} indicator. \\

Facial
& $67$--$129$
& Blendshape velocities: local facial index $67+i$ is the time
derivative of expression weight $i$, for $i=0,\ldots,62$. \\
\midrule
Head, body, left hand, right hand
& $0$--$2$
& Position $(x,y,z)$. \\

Head, body, left hand, right hand
& $3$--$6$
& Orientation quaternion $(x,y,z,w)$. \\

Head, body, left hand, right hand
& $7$--$9$
& Linear velocity $(v_x,v_y,v_z)$. \\

Head, body, left hand, right hand
& $10$--$12$
& Angular velocity $(\omega_x,\omega_y,\omega_z)$. \\
\midrule
Left/right finger
& $0$; $1$
& Hand confidence and hand scale, respectively. \\

Left/right finger
& $2$--$8$
& Pointer pose: position $(x,y,z)$ followed by quaternion $(x,y,z,w)$. \\

Left/right finger
& $9$--$15$
& Root pose: position $(x,y,z)$ followed by quaternion $(x,y,z,w)$. \\

Left/right finger
& $16+4b$--$19+4b$, $b=0,\ldots,25$
& Quaternion $(x,y,z,w)$ for \texttt{boneRotations[b]};
the source-array order is preserved. \\

Left/right finger
& $120$--$124$
& The five \texttt{fingerConfidences} entries, in stored order. \\

Left/right finger
& $125$--$129$
& The five \texttt{pinchStrength} entries, in stored order. \\

Left/right finger
& $130$
& Hand-scale velocity. \\

Left/right finger
& $131$--$136$
& Pointer linear velocity $(v_x,v_y,v_z)$ followed by angular velocity
$(\omega_x,\omega_y,\omega_z)$. \\

Left/right finger
& $137$--$142$
& Root linear velocity followed by angular velocity, in the same order. \\

Left/right finger
& $143+3b$--$145+3b$, $b=0,\ldots,25$
& Angular velocity $(\omega_x,\omega_y,\omega_z)$ of
\texttt{boneRotations[b]}. \\

Left/right finger
& $221$--$225$
& Time derivatives of the five pinch strengths. \\

Left/right finger
& \texttt{flag\_0}--\texttt{flag\_7}
& Bits $0$--$7$ of the recorded hand-status bitmask, decoded as
separate binary indicators. \\

Left/right finger
& \texttt{flag\_8}--\texttt{flag\_12}
& Bits $0$--$4$ of the recorded pinch bitmask, decoded as separate
binary indicators. \\
\midrule
Every modality
& \texttt{present}
& One inside the temporally observed/interpolated support of that
modality and zero elsewhere. \\
\bottomrule
\end{tabular}
\caption{Zero-based local feature mappings within each modality.
Unless a row explicitly names a flag or presence channel, its entries are
indices $i$ in \texttt{M.feature\_i}.
For rows containing two index ranges, the first and second ranges refer to
the left and right eye, respectively.
Finger bone and finger-level arrays retain the order in which they are
stored in the source events.}
\label{tab:feature-mappings}
\end{table*}
\begin{table*}[!t]
\centering
\small
\begin{tabular}{@{}rlrl@{}}
\toprule
$i$ & Blendshape & $i$ & Blendshape \\
\midrule
0 & \texttt{Brow\_Lowerer\_L}
& 32 & \texttt{Lip\_Corner\_Puller\_L}$^{\dagger}$ \\
1 & \texttt{Brow\_Lowerer\_R}
& 33 & \texttt{Lip\_Corner\_Puller\_R}$^{\dagger}$ \\
2 & \texttt{Cheek\_Puff\_L}$^{\dagger}$
& 34 & \texttt{Lip\_Funneler\_LB}$^{\dagger}$ \\
3 & \texttt{Cheek\_Puff\_R}$^{\dagger}$
& 35 & \texttt{Lip\_Funneler\_LT}$^{\dagger}$ \\
4 & \texttt{Cheek\_Raiser\_L}$^{\dagger}$
& 36 & \texttt{Lip\_Funneler\_RB}$^{\dagger}$ \\
5 & \texttt{Cheek\_Raiser\_R}$^{\dagger}$
& 37 & \texttt{Lip\_Funneler\_RT}$^{\dagger}$ \\
6 & \texttt{Cheek\_Suck\_L}$^{\dagger}$
& 38 & \texttt{Lip\_Pucker\_L}$^{\dagger}$ \\
7 & \texttt{Cheek\_Suck\_R}$^{\dagger}$
& 39 & \texttt{Lip\_Pucker\_R}$^{\dagger}$ \\
8 & \texttt{Chin\_Raiser\_B}$^{\dagger}$
& 40 & \texttt{Lip\_Stretcher\_L}$^{\dagger}$ \\
9 & \texttt{Chin\_Raiser\_T}$^{\dagger}$
& 41 & \texttt{Lip\_Stretcher\_R}$^{\dagger}$ \\
10 & \texttt{Dimpler\_L}
& 42 & \texttt{Lip\_Suck\_LB}$^{\dagger}$ \\
11 & \texttt{Dimpler\_R}
& 43 & \texttt{Lip\_Suck\_LT}$^{\dagger}$ \\
12 & \texttt{Eyes\_Closed\_L}
& 44 & \texttt{Lip\_Suck\_RB}$^{\dagger}$ \\
13 & \texttt{Eyes\_Closed\_R}
& 45 & \texttt{Lip\_Suck\_RT}$^{\dagger}$ \\
14 & \texttt{Eyes\_Look\_Down\_L}
& 46 & \texttt{Lip\_Pressor\_L}$^{\dagger}$ \\
15 & \texttt{Eyes\_Look\_Down\_R}
& 47 & \texttt{Lip\_Pressor\_R}$^{\dagger}$ \\
16 & \texttt{Eyes\_Look\_Left\_L}
& 48 & \texttt{Lip\_Tightener\_L}$^{\dagger}$ \\
17 & \texttt{Eyes\_Look\_Left\_R}
& 49 & \texttt{Lip\_Tightener\_R}$^{\dagger}$ \\
18 & \texttt{Eyes\_Look\_Right\_L}
& 50 & \texttt{Lips\_Toward}$^{\dagger}$ \\
19 & \texttt{Eyes\_Look\_Right\_R}
& 51 & \texttt{Lower\_Lip\_Depressor\_L}$^{\dagger}$ \\
20 & \texttt{Eyes\_Look\_Up\_L}
& 52 & \texttt{Lower\_Lip\_Depressor\_R}$^{\dagger}$ \\
21 & \texttt{Eyes\_Look\_Up\_R}
& 53 & \texttt{Mouth\_Left}$^{\dagger}$ \\
22 & \texttt{Inner\_Brow\_Raiser\_L}
& 54 & \texttt{Mouth\_Right}$^{\dagger}$ \\
23 & \texttt{Inner\_Brow\_Raiser\_R}
& 55 & \texttt{Nose\_Wrinkler\_L} \\
24 & \texttt{Jaw\_Drop}$^{\dagger}$
& 56 & \texttt{Nose\_Wrinkler\_R} \\
25 & \texttt{Jaw\_Sideways\_Left}$^{\dagger}$
& 57 & \texttt{Outer\_Brow\_Raiser\_L} \\
26 & \texttt{Jaw\_Sideways\_Right}$^{\dagger}$
& 58 & \texttt{Outer\_Brow\_Raiser\_R} \\
27 & \texttt{Jaw\_Thrust}$^{\dagger}$
& 59 & \texttt{Upper\_Lid\_Raiser\_L} \\
28 & \texttt{Lid\_Tightener\_L}
& 60 & \texttt{Upper\_Lid\_Raiser\_R} \\
29 & \texttt{Lid\_Tightener\_R}
& 61 & \texttt{Upper\_Lip\_Raiser\_L}$^{\dagger}$ \\
30 & \texttt{Lip\_Corner\_Depressor\_L}$^{\dagger}$
& 62 & \texttt{Upper\_Lip\_Raiser\_R}$^{\dagger}$ \\
31 & \texttt{Lip\_Corner\_Depressor\_R}$^{\dagger}$
& & \\
\bottomrule
\end{tabular}
\caption{Mapping of the 63 facial expression-weight indices.
For every listed index $i$, the facial channel with local index $67+i$
contains the corresponding blendshape velocity.
A dagger marks expression weights included in the targeted lower-face
ablation.}
\label{tab:facial-feature-mapping}
\end{table*}
%
\begin{table*}[!t]
\centering
\footnotesize
\setlength{\tabcolsep}{2.6pt}
\begin{tabular}{l rrrr r rrrr r}
\toprule
 & \multicolumn{5}{c}{AUROC} & \multicolumn{5}{c}{Macro-$F_1$} \\
\cmidrule(lr){2-6}\cmidrule(lr){7-11}
Model & $\pm$0.1 & $\pm$0.5 & $\pm$1 & $\pm$2.5 & avg. & $\pm$0.1 & $\pm$0.5 & $\pm$1 & $\pm$2.5 & avg. \\
\midrule
\textsc{RocketPFN} & .680 & \textbf{.748} & .734 & \textbf{.750} & \textbf{.728} & .627 & .686 & .667 & .679 & .665 \\
 & {\tiny$\pm$.033} & {\tiny$\pm$.042} & {\tiny$\pm$.042} & {\tiny$\pm$.045} & {\tiny$\pm$.039} & {\tiny$\pm$.020} & {\tiny$\pm$.036} & {\tiny$\pm$.034} & {\tiny$\pm$.042} & {\tiny$\pm$.031} \\
\addlinespace[1pt]
\textsc{HIVE-COTE~2.0} & .665 & .741 & \textbf{.748} & .737 & .723 & .610 & \textbf{.687} & \textbf{.686} & \textbf{.682} & \textbf{.666} \\
 & {\tiny$\pm$.030} & {\tiny$\pm$.031} & {\tiny$\pm$.037} & {\tiny$\pm$.038} & {\tiny$\pm$.032} & {\tiny$\pm$.025} & {\tiny$\pm$.030} & {\tiny$\pm$.029} & {\tiny$\pm$.029} & {\tiny$\pm$.025} \\
\addlinespace[1pt]
\textsc{DrCIF} & .681 & .731 & .741 & .737 & .722 & .629 & .664 & .674 & .669 & .659 \\
 & {\tiny$\pm$.031} & {\tiny$\pm$.034} & {\tiny$\pm$.041} & {\tiny$\pm$.045} & {\tiny$\pm$.036} & {\tiny$\pm$.030} & {\tiny$\pm$.033} & {\tiny$\pm$.032} & {\tiny$\pm$.037} & {\tiny$\pm$.032} \\
\addlinespace[1pt]
\textsc{MultiRocket+Hydra} & .675 & .710 & .721 & .724 & .707 & .622 & .665 & .669 & .660 & .654 \\
 & {\tiny$\pm$.038} & {\tiny$\pm$.038} & {\tiny$\pm$.034} & {\tiny$\pm$.038} & {\tiny$\pm$.033} & {\tiny$\pm$.030} & {\tiny$\pm$.028} & {\tiny$\pm$.027} & {\tiny$\pm$.031} & {\tiny$\pm$.025} \\
\addlinespace[1pt]
\textsc{MASHT} & \textbf{.688} & .697 & .697 & .699 & .695 & \textbf{.638} & .643 & .638 & .631 & .638 \\
 & {\tiny$\pm$.038} & {\tiny$\pm$.052} & {\tiny$\pm$.050} & {\tiny$\pm$.041} & {\tiny$\pm$.042} & {\tiny$\pm$.031} & {\tiny$\pm$.044} & {\tiny$\pm$.040} & {\tiny$\pm$.036} & {\tiny$\pm$.036} \\
\addlinespace[1pt]
\textsc{SelFRocket} & .672 & .708 & .697 & .686 & .691 & .621 & .652 & .649 & .633 & .639 \\
 & {\tiny$\pm$.036} & {\tiny$\pm$.040} & {\tiny$\pm$.032} & {\tiny$\pm$.039} & {\tiny$\pm$.033} & {\tiny$\pm$.022} & {\tiny$\pm$.027} & {\tiny$\pm$.024} & {\tiny$\pm$.034} & {\tiny$\pm$.023} \\
\addlinespace[1pt]
\textsc{MiniRocket} & .672 & .702 & .707 & .681 & .690 & .628 & .643 & .654 & .628 & .638 \\
 & {\tiny$\pm$.030} & {\tiny$\pm$.036} & {\tiny$\pm$.033} & {\tiny$\pm$.042} & {\tiny$\pm$.031} & {\tiny$\pm$.026} & {\tiny$\pm$.030} & {\tiny$\pm$.024} & {\tiny$\pm$.039} & {\tiny$\pm$.027} \\
\addlinespace[1pt]
Adaptive \textsc{RocketPFN} & .665 & .696 & .697 & .698 & .689 & .613 & .634 & .646 & .628 & .630 \\
 & {\tiny$\pm$.035} & {\tiny$\pm$.041} & {\tiny$\pm$.044} & {\tiny$\pm$.041} & {\tiny$\pm$.038} & {\tiny$\pm$.028} & {\tiny$\pm$.031} & {\tiny$\pm$.037} & {\tiny$\pm$.030} & {\tiny$\pm$.029} \\
\addlinespace[1pt]
\textsc{WEASEL~2.0} & .655 & .676 & .669 & .681 & .670 & .601 & .623 & .621 & .637 & .621 \\
 & {\tiny$\pm$.023} & {\tiny$\pm$.042} & {\tiny$\pm$.038} & {\tiny$\pm$.040} & {\tiny$\pm$.032} & {\tiny$\pm$.021} & {\tiny$\pm$.033} & {\tiny$\pm$.027} & {\tiny$\pm$.038} & {\tiny$\pm$.025} \\
\addlinespace[1pt]
\textsc{GHTT} & .656 & .676 & .681 & .639 & .663 & .605 & .618 & .632 & .597 & .613 \\
 & {\tiny$\pm$.029} & {\tiny$\pm$.039} & {\tiny$\pm$.030} & {\tiny$\pm$.031} & {\tiny$\pm$.024} & {\tiny$\pm$.033} & {\tiny$\pm$.032} & {\tiny$\pm$.020} & {\tiny$\pm$.026} & {\tiny$\pm$.021} \\
\addlinespace[1pt]
Sparse \textsc{MultiRocket} & .631 & .683 & .664 & .658 & .659 & .604 & .604 & .627 & .616 & .613 \\
 & {\tiny$\pm$.037} & {\tiny$\pm$.037} & {\tiny$\pm$.032} & {\tiny$\pm$.044} & {\tiny$\pm$.031} & {\tiny$\pm$.032} & {\tiny$\pm$.042} & {\tiny$\pm$.024} & {\tiny$\pm$.034} & {\tiny$\pm$.023} \\
\addlinespace[1pt]
Inception-\textsc{TCN} & .669 & .652 & .638 & .663 & .656 & .611 & .611 & .592 & .599 & .603 \\
 & {\tiny$\pm$.042} & {\tiny$\pm$.040} & {\tiny$\pm$.038} & {\tiny$\pm$.036} & {\tiny$\pm$.035} & {\tiny$\pm$.031} & {\tiny$\pm$.027} & {\tiny$\pm$.025} & {\tiny$\pm$.029} & {\tiny$\pm$.021} \\
\addlinespace[1pt]
Compact Fusion \textsc{TCN} & .654 & .639 & .624 & .632 & .637 & .600 & .584 & .578 & .585 & .587 \\
 & {\tiny$\pm$.032} & {\tiny$\pm$.037} & {\tiny$\pm$.036} & {\tiny$\pm$.040} & {\tiny$\pm$.036} & {\tiny$\pm$.043} & {\tiny$\pm$.034} & {\tiny$\pm$.035} & {\tiny$\pm$.029} & {\tiny$\pm$.034} \\
\addlinespace[1pt]
\textsc{T2M-GPT~v2} & .651 & .633 & .641 & .618 & .636 & .608 & .572 & .581 & .574 & .584 \\
 & {\tiny$\pm$.032} & {\tiny$\pm$.050} & {\tiny$\pm$.037} & {\tiny$\pm$.044} & {\tiny$\pm$.037} & {\tiny$\pm$.032} & {\tiny$\pm$.043} & {\tiny$\pm$.029} & {\tiny$\pm$.031} & {\tiny$\pm$.027} \\
\addlinespace[1pt]
Event Transformer & .636 & .625 & .618 & .635 & .628 & .608 & .598 & .571 & .581 & .590 \\
 & {\tiny$\pm$.045} & {\tiny$\pm$.051} & {\tiny$\pm$.044} & {\tiny$\pm$.039} & {\tiny$\pm$.043} & {\tiny$\pm$.028} & {\tiny$\pm$.031} & {\tiny$\pm$.027} & {\tiny$\pm$.029} & {\tiny$\pm$.028} \\
\addlinespace[1pt]
\textsc{MotionGPT} & .601 & .638 & .622 & .629 & .622 & .552 & .589 & .573 & .556 & .568 \\
 & {\tiny$\pm$.043} & {\tiny$\pm$.040} & {\tiny$\pm$.039} & {\tiny$\pm$.039} & {\tiny$\pm$.037} & {\tiny$\pm$.048} & {\tiny$\pm$.036} & {\tiny$\pm$.033} & {\tiny$\pm$.045} & {\tiny$\pm$.034} \\
\addlinespace[1pt]
\textsc{T2M-GPT} & .598 & .615 & .626 & .618 & .614 & .561 & .564 & .585 & .566 & .569 \\
 & {\tiny$\pm$.035} & {\tiny$\pm$.035} & {\tiny$\pm$.040} & {\tiny$\pm$.036} & {\tiny$\pm$.034} & {\tiny$\pm$.038} & {\tiny$\pm$.027} & {\tiny$\pm$.033} & {\tiny$\pm$.025} & {\tiny$\pm$.027} \\
\addlinespace[1pt]
\textsc{MrSQM} & .552 & .600 & .637 & .643 & .608 & .531 & .578 & .585 & .599 & .573 \\
 & {\tiny$\pm$.027} & {\tiny$\pm$.019} & {\tiny$\pm$.025} & {\tiny$\pm$.031} & {\tiny$\pm$.016} & {\tiny$\pm$.025} & {\tiny$\pm$.021} & {\tiny$\pm$.021} & {\tiny$\pm$.026} & {\tiny$\pm$.014} \\
\addlinespace[1pt]
\textsc{Castor} & .582 & .598 & .592 & .589 & .590 & .564 & .578 & .571 & .563 & .569 \\
 & {\tiny$\pm$.019} & {\tiny$\pm$.026} & {\tiny$\pm$.035} & {\tiny$\pm$.026} & {\tiny$\pm$.017} & {\tiny$\pm$.017} & {\tiny$\pm$.031} & {\tiny$\pm$.028} & {\tiny$\pm$.022} & {\tiny$\pm$.015} \\
\addlinespace[1pt]
\textsc{MotionGPT-3} & .481 & .485 & .480 & .475 & .480 & .484 & .486 & .484 & .482 & .484 \\
 & {\tiny$\pm$.028} & {\tiny$\pm$.025} & {\tiny$\pm$.025} & {\tiny$\pm$.028} & {\tiny$\pm$.025} & {\tiny$\pm$.021} & {\tiny$\pm$.022} & {\tiny$\pm$.021} & {\tiny$\pm$.021} & {\tiny$\pm$.020} \\
\bottomrule
\end{tabular}
\caption{Cue-centered discrimination from speaker-side multimodal events: mean test performance over 10 cross-validation folds, for 20 time-series classifiers $\times$ 4 temporal windows (s around word onset), with the half-width of the bootstrap 95\% confidence interval below each value.}
\label{tab:model-comparison}
\end{table*}

\begin{table}[t]
\centering
\small
\setlength{\tabcolsep}{2pt}
\begin{tabular}{@{}rlccr@{\;}l@{}}
\toprule
\multicolumn{2}{@{}l}{} & \multicolumn{2}{c}{AUROC} & \multicolumn{2}{c@{}}{} \\
\cmidrule(l{2pt}r{2pt}){3-4}
\# & Model & Val. & Test & \multicolumn{2}{c@{}}{Test \#} \\
\midrule
1 & RocketPFN & .728 & .728 & 1 & \phantom{$\uparrow$} \\
2 & HIVE-COTE~2 & .723 & .723 & 2 & \phantom{$\uparrow$} \\
3 & DrCIF & .722 & .722 & 3 & \phantom{$\uparrow$} \\
4 & MultiRocket & .708 & .707 & 4 & \phantom{$\uparrow$} \\
5 & MASHT & .695 & .695 & 5 & \phantom{$\uparrow$} \\
6 & SelfRocket & .695 & .691 & 6 & \phantom{$\uparrow$} \\
7 & MiniRocket & .689 & .690 & 7 & \phantom{$\uparrow$} \\
8 & AdaptiveRocketPFN & .688 & .689 & 8 & \phantom{$\uparrow$} \\
9 & WEASEL~2 & .678 & .670 & 9 & \phantom{$\uparrow$} \\
\cmidrule(l{2pt}r{2pt}){1-6}
10 & \textbf{Inception-TCN} & .669 & .656 & 12 & $\downarrow$2 \\
11 & \textbf{GHTT} & .667 & .663 & 10 & $\uparrow$1 \\
12 & \textbf{T2M-GPT~2} & .663 & .636 & 14 & $\downarrow$2 \\
13 & \textbf{SparseMultiRocket-Hydra} & .661 & .659 & 11 & $\uparrow$2 \\
14 & \textbf{EventTransformer} & .654 & .628 & 15 & $\downarrow$1 \\
15 & \textbf{CompactFusion-TCN} & .643 & .637 & 13 & $\uparrow$2 \\
16 & \textbf{T2M-GPT} & .639 & .614 & 17 & $\downarrow$1 \\
17 & \textbf{MotionGPT} & .635 & .622 & 16 & $\uparrow$1 \\
\cmidrule(l{2pt}r{2pt}){1-6}
18 & MrSQM & .604 & .608 & 18 & \phantom{$\uparrow$} \\
19 & CASTOR & .596 & .590 & 19 & \phantom{$\uparrow$} \\
20 & MotionGPT-3 & .506 & .480 & 20 & \phantom{$\uparrow$} \\
\bottomrule
\end{tabular}
\caption{Model ranking under validation and test AUROC, averaged over the four symmetric windows and ordered by validation rank. The two orderings agree closely (Spearman $\rho=.985$, Kendall $\tau=.937$; 6 of 190 pairs discordant). Ranks 1--9 and 18--20 are identical across splits; only the eight models in the middle block (bold) change rank, each by at most two positions and only between models separated by $\le.024$ AUROC. Arrows give the rank change from validation to test.}
\label{tab:model-ranking}
\end{table}

\begin{table*}[t]
\centering
\setlength{\tabcolsep}{1.55pt}
\footnotesize
\begin{tabular}{ll rrrrrrrr rrrrrrrr r}
\toprule
& & \multicolumn{8}{c}{before cue onset $[-t,0]$} & \multicolumn{8}{c}{after cue onset $[0,t]$} & \\
\cmidrule(lr){3-10}\cmidrule(lr){11-18}
source & metric & $-$5 & $-$2.5 & $-$2 & $-$1.5 & $-$1 & $-$0.5 & $-$0.25 & $-$0.1 & 0.1 & 0.25 & 0.5 & 1 & 1.5 & 2 & 2.5 & 5 & avg. \\
\midrule
speaker & AUROC & .734 & \textbf{.742} & .734 & .725 & .718 & .705 & .690 & .660 & .678 & .696 & .715 & .717 & .728 & .727 & .723 & .698 & .712 \\
 & & {\tiny$\pm$.045} & {\tiny$\pm$.036} & {\tiny$\pm$.034} & {\tiny$\pm$.036} & {\tiny$\pm$.032} & {\tiny$\pm$.038} & {\tiny$\pm$.037} & {\tiny$\pm$.040} & {\tiny$\pm$.035} & {\tiny$\pm$.040} & {\tiny$\pm$.039} & {\tiny$\pm$.040} & {\tiny$\pm$.043} & {\tiny$\pm$.044} & {\tiny$\pm$.043} & {\tiny$\pm$.036} & \\
 & macro-$F_1$ & .657 & .660 & .648 & .651 & .650 & .650 & .633 & .618 & .630 & .636 & .655 & .647 & .655 & \textbf{.668} & .650 & .642 & .647 \\
 & & {\tiny$\pm$.032} & {\tiny$\pm$.028} & {\tiny$\pm$.031} & {\tiny$\pm$.029} & {\tiny$\pm$.028} & {\tiny$\pm$.029} & {\tiny$\pm$.028} & {\tiny$\pm$.033} & {\tiny$\pm$.032} & {\tiny$\pm$.030} & {\tiny$\pm$.028} & {\tiny$\pm$.035} & {\tiny$\pm$.033} & {\tiny$\pm$.035} & {\tiny$\pm$.034} & {\tiny$\pm$.029} & \\
\addlinespace[2pt]
listener & AUROC & .616 & .617 & .621 & .619 & .607 & .611 & .612 & .596 & .607 & .616 & .600 & .606 & .617 & .619 & .617 & \textbf{.634} & .614 \\
 & & {\tiny$\pm$.033} & {\tiny$\pm$.036} & {\tiny$\pm$.041} & {\tiny$\pm$.034} & {\tiny$\pm$.033} & {\tiny$\pm$.029} & {\tiny$\pm$.034} & {\tiny$\pm$.035} & {\tiny$\pm$.033} & {\tiny$\pm$.028} & {\tiny$\pm$.028} & {\tiny$\pm$.036} & {\tiny$\pm$.034} & {\tiny$\pm$.032} & {\tiny$\pm$.038} & {\tiny$\pm$.040} & \\
 & macro-$F_1$ & .579 & \textbf{.593} & .588 & .591 & .572 & .585 & .577 & .567 & .569 & .577 & .570 & .576 & .579 & .583 & .577 & .584 & .579 \\
 & & {\tiny$\pm$.032} & {\tiny$\pm$.031} & {\tiny$\pm$.033} & {\tiny$\pm$.029} & {\tiny$\pm$.029} & {\tiny$\pm$.022} & {\tiny$\pm$.026} & {\tiny$\pm$.032} & {\tiny$\pm$.027} & {\tiny$\pm$.021} & {\tiny$\pm$.021} & {\tiny$\pm$.023} & {\tiny$\pm$.032} & {\tiny$\pm$.029} & {\tiny$\pm$.026} & {\tiny$\pm$.031} & \\
\addlinespace[2pt]
both & AUROC & .509 & .543 & .551 & .527 & .561 & .572 & .602 & .583 & .585 & .583 & .584 & .598 & .584 & \textbf{.610} & .602 & .602 & .575 \\
 & & {\tiny$\pm$.029} & {\tiny$\pm$.024} & {\tiny$\pm$.025} & {\tiny$\pm$.034} & {\tiny$\pm$.027} & {\tiny$\pm$.022} & {\tiny$\pm$.022} & {\tiny$\pm$.027} & {\tiny$\pm$.022} & {\tiny$\pm$.025} & {\tiny$\pm$.032} & {\tiny$\pm$.041} & {\tiny$\pm$.033} & {\tiny$\pm$.029} & {\tiny$\pm$.026} & {\tiny$\pm$.034} & \\
 & macro-$F_1$ & .494 & .533 & .534 & .501 & .531 & .546 & .562 & .550 & .546 & .542 & .544 & .557 & .542 & \textbf{.566} & .552 & .565 & .542 \\
 & & {\tiny$\pm$.025} & {\tiny$\pm$.023} & {\tiny$\pm$.024} & {\tiny$\pm$.026} & {\tiny$\pm$.025} & {\tiny$\pm$.022} & {\tiny$\pm$.020} & {\tiny$\pm$.022} & {\tiny$\pm$.024} & {\tiny$\pm$.020} & {\tiny$\pm$.023} & {\tiny$\pm$.028} & {\tiny$\pm$.025} & {\tiny$\pm$.021} & {\tiny$\pm$.024} & {\tiny$\pm$.031} & \\
\bottomrule
\end{tabular}
\caption{Cue-centered discrimination performance of the ROCKETPFN probe across one-sided observation windows and interactional sources. Columns give the window extent $t$ in seconds; each value is the mean over ten test folds, with the half-width of its bootstrap 95\% confidence interval below. Best window per row in bold.}
\label{tab:temporal-source}
\end{table*}

We report the configurations used for all comparison models.
Unless stated otherwise, the same configuration is used across temporal
contexts, modality analyses, and interactional-source conditions.
No hyperparameter is selected using a test fold.
\subsection{Shared Input Processing}
Fixed-grid classifiers receive the same modality-by-time representation.
For each window, at most 128 observations per modality are retained using uniform temporal subsampling.
Duplicate timestamps are averaged, and each continuous channel is linearly interpolated at 32 equidistant time points between the left and right window boundaries.
Interpolation is restricted to the interval between the first and last observation of a stream, and positions outside this support are set to zero.
A stream containing only one observation is placed at its nearest grid position.
A separate presence channel indicates the temporal support of each modality.
Numeric event features are standardized using the mean and population standard deviation estimated exclusively from the training split of the current fold.
The 13 finger-status and pinch indicators per finger modality and the modality-presence channels are not standardized.
The resulting input has shape $\mathbf{X}\in\mathbb{R}^{698\times32}$.
Channels are concatenated in the modality order shown in Table~\ref{tab:fixed-grid-channels}.
Within a modality $M$, continuous channels are named \texttt{M.feature\_i}, where $i$ is the zero-based local index defined in Table~\ref{tab:feature-mappings}.
Finger bit indicators are named \texttt{M.flag\_i}, and the support indicator is named \texttt{M.present}.
Positions and linear velocities use $(x,y,z)$ component order.
All orientations are normalized quaternions in $(x,y,z,w)$ order, with the quaternion hemisphere fixed so that $w\geq0$.
Angular velocity is computed from the shortest-path relative quaternion and is represented by its three-axis rotation vector.
Facial and scalar motion features are ordinary first-order rates.
When no preceding observation is available, the corresponding derived motion features are zero.
The targeted lower-face sub-ablation retrains and evaluates the probe with
39 mouth-, lip-, jaw-, cheek-, and chin-related expression weights and their
39 corresponding velocity channels excluded from both training and test
inputs.
These expression weights are marked by daggers in
Table~\ref{tab:facial-feature-mapping}.
Equivalently, the excluded raw blendshape indices are
${2\text{--}9,24\text{--}27,30\text{--}54,61\text{--}62}$, and the
excluded local facial velocity indices are
${69\text{--}76,91\text{--}94,97\text{--}121,128\text{--}129}$.
Thus, 78 channels are excluded in total.
The two facial confidence channels, both facial-validity channels, and the
facial presence channel remain available in this ablation.
For the speaker-only, partner-only, and dyadic conditions, the corresponding actor filters are \texttt{anchor}, \texttt{other}, and \texttt{all}.
In the dyadic fixed-grid representation, simultaneous observations from both participants within the same modality are averaged.
The irregular-event Transformer does not use the fixed-grid representation.
It uses the same local continuous feature mappings in Table~\ref{tab:feature-mappings} and the same limit of 128 observations per modality, but preserves the resulting irregular event sequence and does not add
modality-presence channels.
The actor of each event is encoded as identical to the anchor speaker, different from the anchor speaker, or unknown.
Observations sharing a timestamp and actor are fused into a single temporal token.
Each token is augmented with its time relative to the anchor and its temporal distance to the preceding and following tokens.
Relative times are clipped to $[-10,10]$\,s.
Learned anchor and classification tokens are added to the sequence.
Absolute timestamps, participant and experiment identifiers, database identifiers, lexical information, and audio are excluded from both representations.
\subsection{Cross-Validation, Optimization, and Decisions}
We use ten stratified group partitions, with the recording session as the grouping variable.
For outer fold $k$, partition $k$ is used as the test set, partition $(k+1)\bmod 10$ as the validation set, and the remaining eight partitions as the training set.
Thus, no recording session occurs in more than one split within a fold.
Neural networks use AdamW, binary cross-entropy for discriminative fine-tuning, gradient clipping at $1.0$, and a \textsc{ReduceLROnPlateau} learning-rate schedule unless specified otherwise.
Neural checkpoints are selected by minimum validation loss.
AUROC is the primary evaluation metric and is computed from continuous model scores, making it independent of a decision threshold.
For threshold-dependent metrics, probabilistic classifiers use a threshold of $0.5$ unless stated otherwise.
\textsc{MotionGPT} instead selects its decision threshold on the validation split to maximize macro-$F_1$.
No positive-class weighting is used unless stated otherwise.
The ridge classifiers used by \textsc{MiniRocket}, \textsc{MultiRocket+Hydra}, and \textsc{SelF-Rocket} select $\alpha$ using efficient leave-one-out ridge cross-validation with squared-error scoring.
Their common grid contains 19 half-decade values, $\alpha\in\{10^{-3},10^{-2.5},\ldots,10^{6}\}$.
This selection uses only the outer training split.
\subsection{Random-Convolution Models}
\paragraph{\textsc{MiniRocket}}
We use 10,000 nominal kernels, at most 32 dilations per kernel, and four CPU workers.
Transformed features are scaled without mean centering and classified using ridge regression over the common $\alpha$ grid.
\paragraph{\textsc{MultiRocket+Hydra}}
\textsc{MultiRocket} uses 6,250 nominal kernels, at most 32 dilations per kernel, four pooling features per kernel, the native raw and first-difference representations, and no per-instance normalization.
\textsc{Hydra} uses eight kernels per group, 64 groups, and at most eight input channels per group.
\textsc{MultiRocket} features are scaled without centering, while \textsc{Hydra} counts use sparse square-root standardization.
The two branches are concatenated and classified using ridge regression over the common $\alpha$ grid.
\paragraph{Sparse \textsc{MultiRocket}}
The sparse variant uses the same random transforms as the dense \textsc{MultiRocket+Hydra} model.
Within each outer training fold, ANOVA $F$ scores select $k\in\{5{,}000,10{,}000,15{,}000\}$ features.
Three-fold stratified inner cross-validation, scored by balanced accuracy, jointly selects $k$ and the downstream linear classifier.
Candidate classifiers are ridge regression with $\alpha\in\{1,10,100,10^3,10^4\}$, $L_2$ logistic regression with $C\in\{10^{-3},10^{-2},10^{-1},1,10\}$, elastic-net logistic SGD with $\alpha\in\{10^{-3},10^{-2},10^{-1}\}$ and $l_1$ ratio $\in\{0.2,0.5,0.8\}$, or a linear SVM with $C\in\{10^{-3},10^{-2},10^{-1},1\}$.
The iteration limit is 2,000 and the convergence tolerance is $10^{-3}$.
\paragraph{\textsc{SelF-Rocket}}
We use 10,000 nominal kernels, at most 32 dilations per kernel, and no per-instance normalization.
We evaluate all 15 combinations of the base, first-difference, and mixed representations with PPV, zero-crossing, MPV, MIPV, and LSPV pooling.
Candidate selection uses two folds repeated ten times, at most 500 cases and 2,500 features per candidate, and ten logarithmically spaced ridge penalties from $10^{-3}$ to $10^3$.
The candidate with the highest median validation accuracy is retained if it ranks among the top five in at least 90\% of runs.
Otherwise, PPV-MIX is used for the present sequence length ($32<512$).
The final ridge classifier uses the common $\alpha$ grid.
\subsection{Feature-, Dictionary-, Shapelet-, and Interval-Based Models}
\paragraph{\textsc{Castor}}
We use 128 groups with 16 shapelets per group, shapelet length 9, Euclidean distance, normalization probability $0.5$, and lower and upper quantiles of $0.01$ and $0.20$.
Soft minimum and soft threshold are enabled, while soft maximum is disabled.
Class labels are available during shapelet sampling (\path{ignore_y=false}).
Half of the groups operate on the raw series and half on first differences.
Sparse features are square-root transformed and standardized with zero-frequency exponent 4.
Classification uses ridge regression with $\alpha\in\{0.01,1,10\}$.
\paragraph{\textsc{Weasel 2.0}}
The minimum window length is 4, window normalization is disabled, and word lengths 7 and 8 are used.
Both raw and first-difference variants are included.
Chi-squared top-$k$ selection retains at most 30,000 features.
The dataset-size rule yields 100 ensemble configurations per fold.
At most 32 channels are selected using training data only.
Ridge classification uses ten logarithmically spaced penalties from $10^{-1}$ to $10^5$.
\paragraph{\textsc{MrSQM}}
We use the random-search strategy with 500 retained and 2,000 preselected features per representation.
The representation uses zero SAX and five SFA representations.
SFA normalization and first differences are enabled.
At most eight channels are selected using the training split.
Classification uses class-balanced logistic regression with $C=1$, the Newton--CG solver, and at most 1,000 iterations.
\paragraph{\textsc{DrCIF}}
The standalone \textsc{DrCIF} classifier uses 200 trees.
Interval counts are $(4,\texttt{sqrt-div})$, with minimum interval length 3 and maximum interval length $0.5$ of the series.
Ten attributes are sampled per interval.
\textsc{Catch22} features are disabled.
No time contract is used, and numerically near-constant intervals are stabilized.

\paragraph{\textsc{HIVE-COTE 2.0}}
We use the four standard \textsc{HIVE-COTE 2.0} components with fourth-power CAWPE weighting.
A nominal 360-minute contract is provided per outer fold.
The implementation-specific component configurations are given below.
\begin{itemize}
    \item \textbf{Shapelet Transform Classifier:} 10,000 shapelet samples are used, with unconstrained maximum shapelet count and length, batch size 100, and a contract cap of 10,000 samples.
    Rotation Forest uses 200 trees with a contract cap of 200.

    \item \textbf{DrCIF component:} The internal \textsc{DrCIF} component uses 500 trees, interval counts $(4,\texttt{sqrt-div})$, minimum interval length 3, maximum proportional interval length $0.5$, and ten attributes per interval.
    \textsc{Catch22} is disabled and near-constant interval stabilization is enabled.
    This configuration is distinct from the standalone \textsc{DrCIF} baseline above.

    \item \textbf{Arsenal:} The ROCKET transform uses 2,000 kernels per estimator, producing maximum and PPV features.
    The ensemble contains 25 estimators with a contract cap of 25 estimators.
    The configured maximum-dilation and features-per-kernel arguments are inactive for the aeon ROCKET branch.

    \item \textbf{Temporal Dictionary Ensemble:} We use 250 parameter samples, a maximum ensemble size of 50, a maximum window-length proportion of $1.0$, and a minimum window length of 10.
    Fifty parameter settings are selected randomly, bigram configuration is automatic, the dimension threshold is $0.85$, and at most 20 dimensions are retained.
    The contract cap is 250 parameter samples.
\end{itemize}
\subsection{Random-Convolution Features with Tabular Inference}
\paragraph{\textsc{RocketPFN}}
We use ten independent ROCKET groups with 1,000 kernels each and per-instance normalization.
Each kernel contributes maximum and PPV features, yielding 2,000 features per group.
One TabPFN~v2.5 classifier with eight internal estimators is fitted per group.
Predicted probabilities are averaged across groups.
TabPFN uses \path{fit_preprocessors} mode, automatic device selection, memory-saving and precision settings, one preprocessing worker, and no probability balancing.
\paragraph{\textsc{MASHT}}
The nominal feature budget is selected deterministically from the total number $N$ of examples.
The budget is 10,000 for $N<1{,}000$, 2,000 for $1{,}000\leq N<100{,}000$, and 200 otherwise.
Only split sizes, rather than held-out values or labels, enter this rule.
The balanced datasets considered here fall in the second range and therefore use a nominal budget of 2,000 features.
Half of this budget is assigned to \textsc{MultiRocket} and half to \textsc{Hydra}.
\textsc{MultiRocket} divides its budget equally between raw and first-difference representations, uses four pooling features per kernel, at most 32 dilations, and no per-instance normalization.
\textsc{Hydra} uses eight kernels per group and at most eight channels.
Only the \textsc{Hydra} branch receives sparse square-root scaling.
The concatenated representation is classified by TabPFN~3 with eight estimators and automatic estimator scaling.
\paragraph{Adaptive \textsc{RocketPFN}}
Our adaptive variant constructs raw, first-difference, second-difference, five-point smoothed, high-pass, and local-normalized views with $\epsilon=10^{-6}$.
Each view is transformed using \textsc{MultiRocket} banks with dilation caps 1 and 32, 625 kernels per bank, four pooling features per kernel, and no per-instance normalization.
The candidate bank additionally contains \textsc{Hydra} with eight kernels per group, 128 groups, and at most eight channels.
It further contains at most 512 random dilated shapelets of lengths 7, 9, and 11, with normalization probability $0.8$ and similarity threshold $0.5$.
Morphology, dynamics, and prototype experts each select 500 features using three-fold, two-repeat inner selection.
At most 12 feature families are active per expert, with at least eight features per active family.
The selection-pool multiplier is 3, the allocation temperature is $0.5$, and the redundancy threshold is $0.98$, estimated from at most 256 cases.
Three-fold out-of-fold predictions from two-estimator TabPFN models determine nonnegative ensemble weights by regularized log loss with $L_2=0.05$.
Final experts use TabPFN~3 with eight estimators.
\subsection{Neural Time-Series and Event Models}
\paragraph{\textsc{Inception-TCN}}
The model contains eight modality-specific branches.
Each branch uses 16 projection channels and two inception blocks with eight channels per branch and kernels of size 3, 5, and 9.
These are followed by depthwise-separable TCN blocks with dilations 1, 2, and 4.
Mask-aware mean and maximum pooling and the observed fraction are concatenated across modalities.
The classifier hidden size is 64, dropout is $0.30$, and normalized relative time is appended.
Training uses batch size 32 and batch size 64 for evaluation.
Models are trained for at most 100 epochs with learning rate $3\times10^{-4}$ and weight decay $10^{-3}$.
Early-stopping patience is 15, and scheduler patience is 5 with factor $0.5$ and minimum learning rate $10^{-6}$.
Mixed precision is enabled.
\paragraph{Compact Fusion-\textsc{TCN}}
Per-modality projection widths are $(8,16,8,8,8,8,24,24)$ for eye, facial, head, body, left/right hand, and left/right finger modalities, respectively.
The fused representation has width 48.
The shared TCN uses kernel size 5, dilations 1, 2, and 4, and two convolutions per block.
Global, pre-anchor, and post-anchor regions are pooled using mean, maximum, and standard deviation together with modality-presence indicators.
The classifier hidden size is 32, dropout is $0.45$, and modality dropout is $0.15$.
Training uses batch size 32 and batch size 64 for evaluation.
Models are trained for at most 100 epochs with learning rate $3\times10^{-4}$ and weight decay $10^{-2}$.
Gaussian input noise with $\sigma=0.02$ and temporal shifts of at most one grid step are applied during training.
Early-stopping patience is 20, and scheduler patience is 6 with factor $0.5$ and minimum learning rate $10^{-6}$.
Mixed precision is enabled.
Results are aggregated over seeds $\{17,42,73\}$.
\paragraph{Irregular-Event Transformer}
The model uses one pre-norm Transformer encoder layer with model width 32, four attention heads, feed-forward width 128, and dropout $0.35$.
Modality encoders and the modality-fusion MLP have hidden width 64.
The time encoder has width 32, and the classifier hidden layer has width 16.
Complete modality observations are masked with probability $0.20$ and reconstructed during ten self-supervised epochs on the outer training split.
Self-supervised pretraining uses learning rate $10^{-4}$ and weight decay $10^{-3}$.
Fine-tuning uses batch size 8 for at most 60 epochs.
Backbone and classifier learning rates are $10^{-4}$ and $3\times10^{-4}$, respectively, with weight decay $10^{-2}$.
The pretrained backbone is frozen for the first three fine-tuning epochs.
Early-stopping patience is 10, and scheduler patience is 3 with factor $0.5$ and minimum learning rate $10^{-6}$.
Mixed precision is enabled.
\subsection{Motion-Language and Hierarchical Motion Models}
All five motion-model adaptations in this subsection use the same normalized fixed-grid input with 32 temporal positions.
\textsc{T2M-GPT} and \textsc{MotionGPT} use the same VQ-VAE tokenizer.
The tokenizer has 128 codes of dimension 64, hidden width 128, two temporal downsampling layers, two residual blocks, dilation growth 3, ReLU activations, and no dropout.
The EMA codebook uses decay $0.99$, codebook $\epsilon=10^{-5}$, and reset threshold 1.
The 32 fixed-grid positions are thereby compressed to eight motion tokens.
The tokenizer is trained for at most 100 epochs using batch size 32 and batch size 64 for evaluation.
Training uses AdamW with learning rate $2\times10^{-4}$ and zero weight decay.
The reconstruction objective is smooth-$L_1$ augmented by velocity and commitment losses with weights $0.5$ and $0.02$, respectively.
The tokenizer checkpoint is selected by validation reconstruction loss with patience 12.
Its learning-rate scheduler uses patience 6, factor $0.5$, and minimum learning rate $10^{-6}$.
Mixed precision is disabled for all motion models.
\paragraph{\textsc{T2M-GPT}}
We use a causal discriminative Transformer with width 64, four attention heads, two layers, feed-forward width 256, dropout $0.30$, a 32-unit classifier, last-token pooling, and token-corruption rate $0.10$.
Before supervised fine-tuning, the backbone undergoes 20 epochs of next-token pretraining with batch size 32, learning rate $3\times10^{-4}$, and weight decay $10^{-2}$.
Fine-tuning uses batch size 32 and batch size 64 for evaluation.
The model is fine-tuned for at most 60 epochs with learning rate $3\times10^{-4}$ and weight decay $10^{-2}$.
Early-stopping patience is 10, and scheduler patience is 4 with factor $0.5$ and minimum learning rate $10^{-6}$.
The fixed decision threshold is $0.5$.
\paragraph{\textsc{T2M-GPT-v2}}
\textsc{T2M-GPT-v2} is our continuous-latent ablation of \textsc{T2M-GPT}; it is not a separate published architecture.
The discrete VQ-VAE is replaced by a variational autoencoder with latent dimension 64, hidden width 128, two temporal downsampling layers, two residual blocks, dilation growth 3, ReLU activations, and no dropout.
The 32 fixed-grid positions are thereby compressed to eight continuous latent vectors.
The VAE is trained for at most 100 epochs using batch size 32 and batch size 64 for evaluation.
Training uses AdamW with learning rate $2\times10^{-4}$ and zero weight decay.
Its objective combines smooth-$L_1$ reconstruction loss, velocity loss with weight $0.5$, and KL divergence with weight $10^{-4}$; free bits are disabled.
The VAE checkpoint is selected by validation loss with early-stopping patience 12 and minimum improvement $10^{-5}$.
Its learning-rate scheduler uses patience 6, factor $0.5$, and minimum learning rate $10^{-6}$.
The causal Transformer has width 64, four attention heads, two layers, feed-forward width 256, dropout $0.30$, a 32-unit classifier, last-token pooling, and latent-corruption rate $0.10$.
The backbone undergoes 20 epochs of next-latent pretraining using smooth-$L_1$ regression, batch size 32, learning rate $3\times10^{-4}$, and weight decay $10^{-2}$.
Fine-tuning is performed for at most 60 epochs with batch size 32, evaluation batch size 64, learning rate $3\times10^{-4}$, and weight decay $10^{-2}$.
Early-stopping patience is 10, and scheduler patience is 4 with factor $0.5$ and minimum learning rate $10^{-6}$.
\paragraph{\textsc{MotionGPT}}
We use a motion-to-text encoder--decoder with model width 64, four attention heads, two encoder layers, two decoder layers, feed-forward width 256, dropout $0.30$, tied word embeddings, maximum sequence length 64, and eight sentinel tokens.
The model is pretrained for 25 epochs on denoising, future-prediction, and in-between completion objectives.
Pretraining uses batch size 32, learning rate $3\times10^{-4}$, and weight decay $10^{-2}$.
Span corruption is $0.25$ with mean span length 2.
Prediction-context and in-between-span fractions are $0.50$ and $0.25$, respectively.
Supervised tuning predicts the label word and otherwise uses the same optimization settings as \textsc{T2M-GPT}.
The answer-token log-odds threshold is selected on the validation split to maximize macro-$F_1$.
\paragraph{\textsc{MotionGPT3}}
We evaluate a continuous-latent classification adaptation of \textsc{MotionGPT3}.
Because no textual descriptions are available, its language branch and cross-modal attention are omitted; the portable continuous representation and diffusion-head mechanism are retained.
The model uses the same continuous VAE and VAE-training configuration as \textsc{T2M-GPT-v2}, producing eight 64-dimensional latent vectors per window.
A non-causal Transformer summarizer with width 64, four attention heads, two layers, feed-forward width 256, and dropout $0.30$ processes these vectors and produces a mean-pooled motion representation.
The class-conditioned diffusion head has hidden width 256, two residual blocks, and dropout $0.10$.
It uses 100 diffusion timesteps and a linear noise schedule from $\beta_1=10^{-4}$ to $\beta_{100}=0.02$.
The denoising objective predicts the added Gaussian noise using mean-squared error.
Training and classification scores average four independently sampled timestep--noise pairs per motion window.
Classification compares the denoising losses obtained under the positive- and negative-class embeddings.
Unlike the \textsc{T2M-GPT} variants, the diffusion classifier receives no separate next-token or next-latent pretraining.
It is trained for at most 60 epochs using batch size 32, evaluation batch size 64, AdamW with learning rate $3\times10^{-4}$ and weight decay $10^{-2}$, and early-stopping patience 10.
The learning-rate scheduler uses patience 4, factor $0.5$, and minimum learning rate $10^{-6}$.
\paragraph{\textsc{G-HTT}}
Our \textsc{G-HTT} adaptation hierarchically models short- and long-span temporal structure.
Each 32-position window is divided into four non-overlapping clips of length eight.
The short-span pose block is a Transformer VAE that encodes the first four positions of each clip, reconstructs those positions, and predicts the remaining four.
Its encoder and two decoders use width 64, four attention heads, two layers, feed-forward width 256, dropout $0.10$, and a 32-dimensional latent bottleneck.
The component-reconstruction and trajectory-prediction losses both have weight 1.
Each is a smooth-$L_1$ objective augmented by a velocity loss with weight $0.5$.
The pose-block KL term has weight $10^{-5}$, and free bits are disabled.
The pose block is trained for at most 100 epochs with batch size 32, evaluation batch size 64, AdamW learning rate $2\times10^{-4}$, and zero weight decay.
Its early-stopping patience is 12, while scheduler patience is 6 with factor $0.5$ and minimum learning rate $10^{-6}$.
The posterior mean of each clip yields a sequence of four 32-dimensional mid-level representations.
The long-span action block is another Transformer VAE with width 64, four attention heads, two layers, feed-forward width 256, dropout $0.30$, and latent dimension 32.
Its classifier contains one hidden layer of width 32.
The action-block objective combines smooth-$L_1$ reconstruction of the mid-level sequence with weight 1, binary cross-entropy with weight $0.1$, and KL divergence with weight $10^{-5}$.
The action block is trained for at most 60 epochs using batch size 32, evaluation batch size 64, AdamW learning rate $3\times10^{-4}$, and weight decay $10^{-2}$.
Early-stopping patience is 10, and scheduler patience is 4 with factor $0.5$ and minimum learning rate $10^{-6}$.
\subsection{Software}
Experiments use Python~3.12.3, NumPy~2.3.5~\citep{Harris:etal:2020}, scikit-learn~1.8.0~\citep{Pedregosa:etal:2011}, SciPy~1.17.1~\citep{Virtanen:etal:2020}, aeon~1.5.0~\citep{Middlehurst:etal:2024}, Wildboar~1.2.1~\citep{Samsten:etal:2024}, MrSQM~0.0.7~\citep{Nguyen:Ifrim:2023}, PyTorch~2.13.0~\citep{Paszke:etal:2019}, Transformers~4.51.3~\citep{Wolf:etal:2020}, and TabPFN~8.1.0~\citep{Hollmann:etal:2025}.
\end{document}